\documentclass[runningheads]{llncs}
\usepackage{graphicx}

\usepackage{tikz}
\usepackage{comment}
\usepackage{amsmath,amssymb} 
\usepackage{color}
\usepackage{xcolor}
\usepackage[accsupp]{axessibility}  
\usepackage{multirow}
\usepackage{booktabs}

\usepackage[pagebackref,breaklinks,colorlinks,citecolor=black,urlcolor=blue]{hyperref}
\newcommand{\OURS}{4DContrast}
\newcommand{\yjnote}[1]{\textcolor{magenta}{[Yujin:\@ #1]}}

\definecolor{green_fig}{RGB}{112,173,71}
\definecolor{orange_fig}{RGB}{197,90,17}
\definecolor{blue_fig}{RGB}{46,117,182}
\definecolor{darkgreen}{RGB}{65,162,52}
\definecolor{darkblue}{RGB}{66,154,214}
\begin{document}
\pagestyle{headings}
\mainmatter
\def\ECCVSubNumber{4908}  

\title{\OURS{}: Contrastive Learning with Dynamic Correspondences for 3D Scene Understanding}

\titlerunning{\OURS{}}
%

\author{Yujin Chen \and Matthias Nießner \and Angela Dai 
}
\authorrunning{Y. Chen et al.}
%
\institute{Technical University of Munich
\\
\email{\{yujin.chen,niessner,angela.dai\}@tum.de}
}
\maketitle

\begin{figure}[th!]
   \vspace{-0.5cm}
     \centering
        \includegraphics[width=1\textwidth]{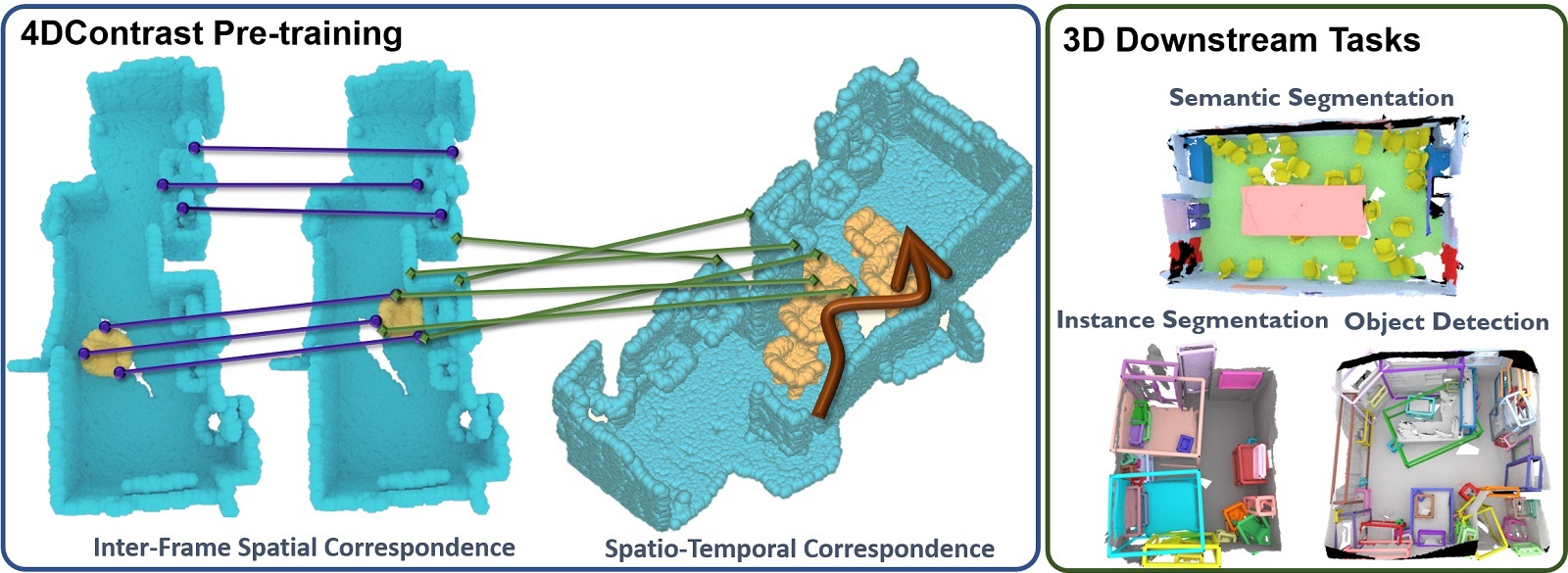}
   \label{fig:teaser}
   \vspace{-0.16in}
       \caption{We propose  \OURS{} to imbue learned 3D representations with 4D priors.
    We introduce a data augmentation scheme to composite synthetic 3D objects with real-world 3D scans to create 4D sequence data with inherent correspondence information.
    We then leverage a combination of 3D-3D, 3D-4D, and 4D-4D constraints within a contrastive learning framework to learn 4D invariance in the 3D representations.
    The learned features can be transferred to improve performance in various downstream 3D scene understanding tasks.}
\end{figure}
\vspace{-1cm}

\begin{abstract}
We present a new approach to instill 4D dynamic object priors into learned 3D representations by unsupervised pre-training.
We observe that dynamic movement of an object through an environment provides important cues about its objectness, and thus propose to imbue learned 3D representations with such dynamic understanding, that can then be effectively transferred to improved performance in downstream 3D semantic scene understanding tasks.
We propose a new data augmentation scheme leveraging synthetic 3D shapes moving in static 3D environments, and employ contrastive learning under 3D-4D constraints that encode 4D invariances into the learned 3D representations.
Experiments demonstrate that our unsupervised representation learning results in improvement in downstream 3D semantic segmentation, object detection, and instance segmentation tasks, and moreover, notably improves performance in data-scarce scenarios. Our results show that our 4D pre-training method improves downstream tasks such as object detection mAP@0.5 by 5.5\%/6.5\% over training from scratch on ScanNet/SUN RGB-D while involving no additional run-time overhead at test time.
\end{abstract}

\section{Introduction}
\label{sec:intro}

3D semantic scene understanding has seen remarkable progress in recent years, in large part driven by advances in deep learning as well as the introduction of large-scale, annotated datasets \cite{chang2017matterport3d,dai2017scannet,geiger2012we}. 
In particular, notable progress has been made to address core 3D scene understanding tasks such as 3D semantic segmentation, object detection, and instance segmentation, which are fundamental to many real-world computer vision applications such as robotics, mixed reality, or autonomous driving.
Such approaches have developed various methods to learn on different 3D scene representations, such as sparse or dense volumetric grids \cite{choy20194d,dai2017scannet,graham20183d}, point clouds \cite{qi2017pointnet,qi2017pointnet++}, meshes \cite{huang2019texturenet}, or multi-view approaches \cite{dai20183dmv,su2015multi}.
Recently, driven by the success of unsupervised representation learning for transfer learning in 2D, 3D scene understanding methods have been augmented with unsupervised 3D pre-training to further improve performance to downstream 3D scene understanding tasks \cite{hou2021exploring,huang2021spatio,xie2020pointcontrast,Zhang_2021_ICCV}.

While such 3D representation learning has focused on feature representations learned from static 3D scenes, we observe that important notions of objectness are given by 4D dynamic observations -- for instance, object segmentations can often be naturally intuited by observing objects moving around an environment without any annotations required, which can be more difficult in a static 3D observation.
We thus propose to leverage this powerful 4D signal in unsupervised pre-training to imbue 4D object priors into learned 3D representations, that can then be effectively transferred to various downstream 3D scene understanding tasks for improved recognition performance.

In this work, we introduce \OURS{} to learn about objectness from both static 3D and dynamic 4D information in learned 3D representations.
We leverage a combination of static 3D scanned scenes and a database of synthetic 3D shapes, and augment the scenes with moving synthetic shapes to generate 4D sequence data with inherent motion correspondence.
We then employ a contrastive learning scheme under both 3D and 4D constraints, correlating local 3D point features with each other as well as with 4D sequence features, thus imbuing learned objectness from dynamic information into the 3D representation learning.

To demonstrate our approach, we pre-train on ScanNet~\cite{dai2017scannet} along with ModelNet~\cite{wu20153d} shapes for unsupervised 3D representation learning.
Experiments on 3D semantic segmentation, object detection, and instance segmentation show that \OURS{} learns effective features that can be transferred to achieve improved performance in various downstream 3D scene understanding tasks.
\OURS{} can also generalize from pre-training on ScanNet and ModelNet to improved performance on SUN RGB-D~\cite{song2015sun}.
Additionally, we show that our learned representations remain robust in limited training data scenarios, consistently improving performance under a various amounts of training data available.

Our main contributions are summarized as follows:
\begin{itemize}
    \item We propose the first method to leverage 4D sequence information and constraints for 3D representation learning, showing transferability of the learned features to the downstream 3D scene understanding tasks of 3D semantic segmentation, object detection, and instance segmentation.
    \item Our new unsupervised pre-training based on constructing 4D sequences from synthetic 3D shapes in real-world, static 3D scenes improves performance across a variety of downstream tasks and different datasets.
\end{itemize}

\vspace{-0.1in}

\section{Related Work}

\noindent \textbf{3D Semantic Scene Understanding.}
Driven by rapid developments in deep learning and the introduction of several large-scale, annotated 3D datasets \cite{chang2017matterport3d,dai2017scannet,geiger2012we}, notable progress has been made in 3D semantic scene understanding, in particular the tasks of 3D semantic segmentation \cite{choy20194d,dai2017scannet,dai20183dmv,graham20183d,hu2021bidirectional,huang2019texturenet,nekrasov2021mix3d,qi2017pointnet++,rozenberszki2022language}, 3D object detection \cite{nie2021rfd,qi2020imvotenet,qi2019deep,xie2020mlcvnet,zhang2020h3dnet}, and 3D instance segmentation \cite{engelmann20203d,han2020occuseg,hou20193d,jiang2020pointgroup,zhang2021point}. 
Many methods have been proposed, largely focusing on learning on various 3D representations, such as sparse or dense volumetric grids \cite{choy20194d,dai2017scannet,graham20183d}, point clouds \cite{jiang2020pointgroup,qi2019deep,qi2017pointnet,qi2017pointnet++}, meshes \cite{huang2019texturenet,schult2020dualconvmesh}, or multi-view hybrid representations \cite{dai20183dmv,kundu2020virtual}.
In particular, approaches leveraging backbones built with sparse convolutions~\cite{choy20194d,graham20183d} have shown strong effectiveness across a variety of 3D scene understanding tasks and datasets.
We propose a new unsupervised pre-training approach to learn 4D priors in learned 3D representations, leveraging sparse convolutional backbones for both 3D and 4D feature extraction.

\noindent \textbf{3D Representation Learning.} 
Inspired by the success of representation learning in 2D, particularly that leveraging instance discrimination with contrastive learning \cite{chen2020simple,chen2020improved,he2020momentum}, recent works have explored unsupervised learning with 3D pretext tasks that can be leveraged for fine-tuning on downstream 3D scene understanding tasks  \cite{chen2021shape,hassani2019unsupervised,hou2021exploring,huang2021spatio,liang2021exploring,rao2021randomrooms,sanghi2020info3d,sauder2019self,wang2021unsupervised,wang2021unsupervisedAAAI,xie2020pointcontrast,Zhang_2021_ICCV}. 
For instance, \cite{hassani2019unsupervised,sauder2019self} learn feature representations from point-based instance discrimination for object classification and segmentation, and \cite{hou2021exploring,xie2020pointcontrast,Zhang_2021_ICCV} extend to more complex 3D scenes by generating correspondences from various different views of scene point clouds.

In particular, given the more scarce data availability of real-world 3D environments, Hou et al.~\cite{hou2021exploring} additionally demonstrate the efficacy of contrastive 3D pretraining for various 3D semantic scene understanding tasks under a variety of limited training data scenarios.
In contrast to these methods that employ 3D-only pretext tasks for representation learning, we propose to learn from 4D sequence data to embed 4D priors into learned 3D representations for more effective transfer to downstream 3D tasks.

Recently, Huang et al.~\cite{huang2021spatio} propose to learn from the inherent sequence data of RGB-D video to incorporate the notion of a temporal sequence.
Constraints are established across pairs of frames in the sequence; however, the sequence data itself represents static scenes without any movement within the scene, limiting the temporal signal that can be learned.
In contrast, we consider 4D sequence data containing object movement through the scene, which can provide additional semantic signal about objectness through an object's motion.
Additionally, Rao et al.~\cite{rao2021randomrooms} propose to learn from 3D scenes that are synthetically generated by randomly placing synthetic CAD models on a rectangular layout.
They employ object-level contrastive learning on object-level features, resulting in improved 3D object detection performance.
We also leverage synthetic CAD models for data augmentation, but we compose them with real-world 3D scan data to generate 4D sequences of objects in motion, and exploit learned 4D features to enhance learned 3D representations, with performance improvement on various downstream 3D scene understanding tasks.

\begin{figure*}[t]
  \centering
   \includegraphics[width=0.99\linewidth]{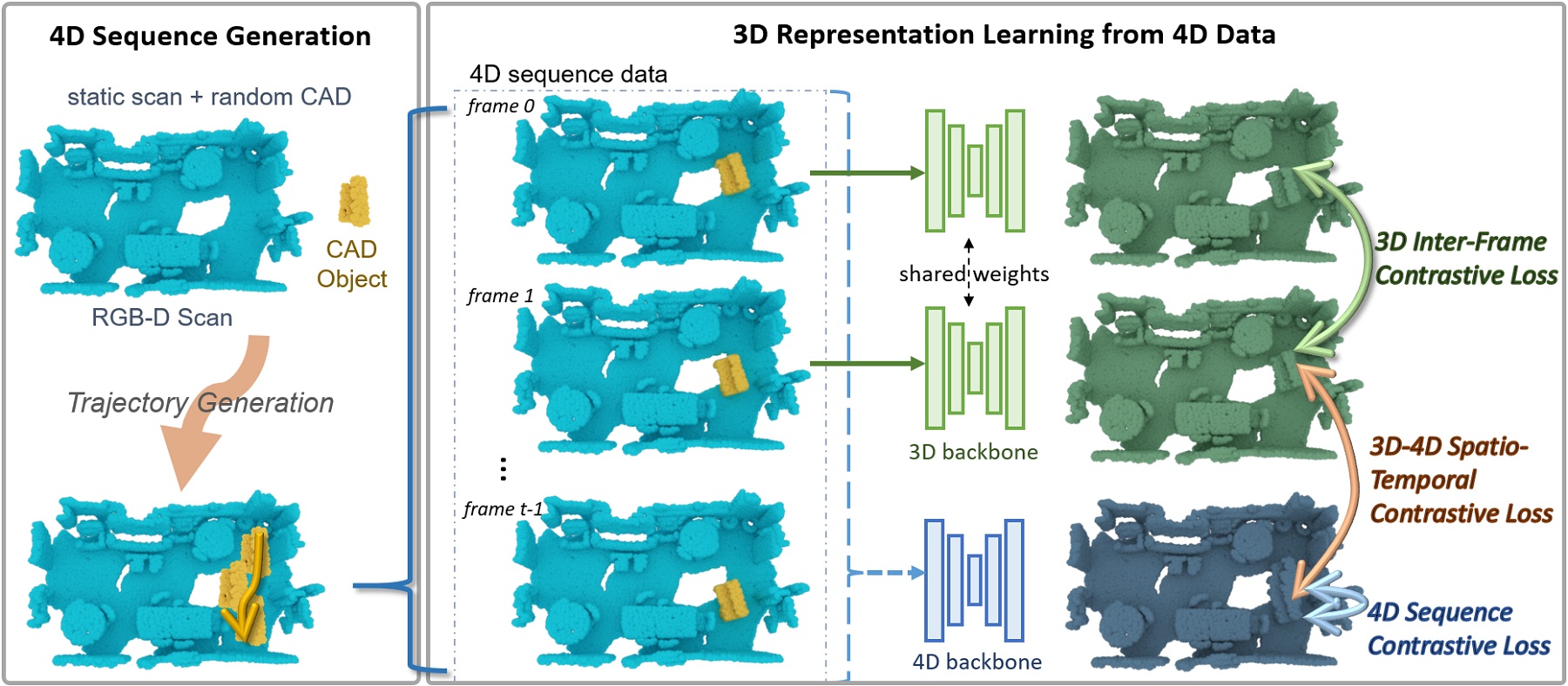}
   \vspace{-0.3cm}
   \caption{Method overview. 
   \OURS{} learns effective 3D feature representations imbued with 4D signal from moving object sequences.
   During pre-training, we augment static 3D scene data with a moving object from a synthetic shape dataset.
   We can then establish dynamic correspondences between the spatio-temporal features learned from the 4D sequence with 3D features of individual static frames.
   We employ contrastive learning under not only 3D geometric correspondence between individual frames, but also with their corresponding 4D counterpart, as well as 4D-4D constraints to anchor the 4D feature learning.
   This enables 4D-invariant representation learning, which we can apply to various downstream 3D scene understanding tasks.
   }
   \label{fig:method}
   \vspace{-5mm}
\end{figure*}

\section{4D Invariant Representation Learning}

\OURS{} presents a new approach to 3D representation learning: our key idea is to employ 4D constraints during pre-training, in order to imbue learned features with 4D invariance from learned objectness from seeing an object in motion.
We consider a dataset of 3D scans $\mathcal{S}=\{S_i\}$ as well as a dataset of synthetic 3D objects $\mathcal{O} = \{O_j\}$, and construct dynamic sequences with inherent correspondence information by moving a synthetic object $O_j$ in a static 3D scan $S_i$.
This enables us to establish 4D correspondences along with 3D-4D correspondence as constraints under a contrastive learning framework for unsupervised pre-training.
An overview of our approach is shown in Figure~\ref{fig:method}.

\subsection{Revisiting SimSiam}
\label{subsec:simsiam}
We first revisit SimSiam~\cite{chen2021exploring}, which introduced a simple yet powerful approach for contrastive 2D representation learning. Inspired by the effectiveness of SimSiam, we build an unsupervised contrastive learning scheme for embedding 4D priors into 3D representations.

SimSiam considers two augmented variants of an image $I$, $I_1$, and $I_2$, which are input to weight-shared encoder network $\Phi_{\textrm{2D}}$ (a 2D convolutional backbone followed by a projection MLP). Then a prediction MLP head $P_{\textrm{2D}}$ transforms the output of one view as $p_1^{\textrm{2D}}=P_{\textrm{2D}}(\Phi_{\textrm{2D}}(I_1))$ to match to the another output $z_2^{\textrm{2D}}=\Phi_{\textrm{2D}}(I_2)$, with minimizing the negative cosine similarity \cite{grill2020bootstrap}:
\begin{equation}
\label{equ:cosine}
\mathcal{D}(p_1^{\textrm{2D}}, z_2^{\textrm{2D}})  = -\frac{p_1^{\textrm{2D}}}{{||p_1^{\textrm{2D}}||}_2}\cdot \frac{z_2^{\textrm{2D}}}{{||z_2^{\textrm{2D}}||}_2}.
\end{equation}
SimSiam also uses a stop-gradient ($SG$) operation that treats $z_2^{\textrm{2D}}$ as a constant during back-propagation,
to prevent collapse during the training, thus modifying Eq.~\ref{equ:cosine} as: $\mathcal{D}(p_1^{\textrm{2D}}, SG(z_2^{\textrm{2D}}))$. A symmetrized loss is defined for the two augmented inputs:
\begin{equation}
\label{equ:sym_cosine}
\mathcal{L^\textrm{2D}}  = \frac{1}{2}\mathcal{D}(p_1^{\textrm{2D}}, SG(z_2^{\textrm{2D}})) +  \frac{1}{2}\mathcal{D}(p_2^{\textrm{2D}}, SG(z_1^{\textrm{2D}})).
\end{equation}

SimSiam has shown to be very effective at learning invariances under various image augmentations, without requiring negative samples or very large batches.
We thus build from this contrastive framework for our 3D-4D constraints, as it allows for our high-dimensional pre-training design.

\begin{figure*}[t]
  \centering
  \vspace{-2mm}
   \includegraphics[width=0.99\linewidth]{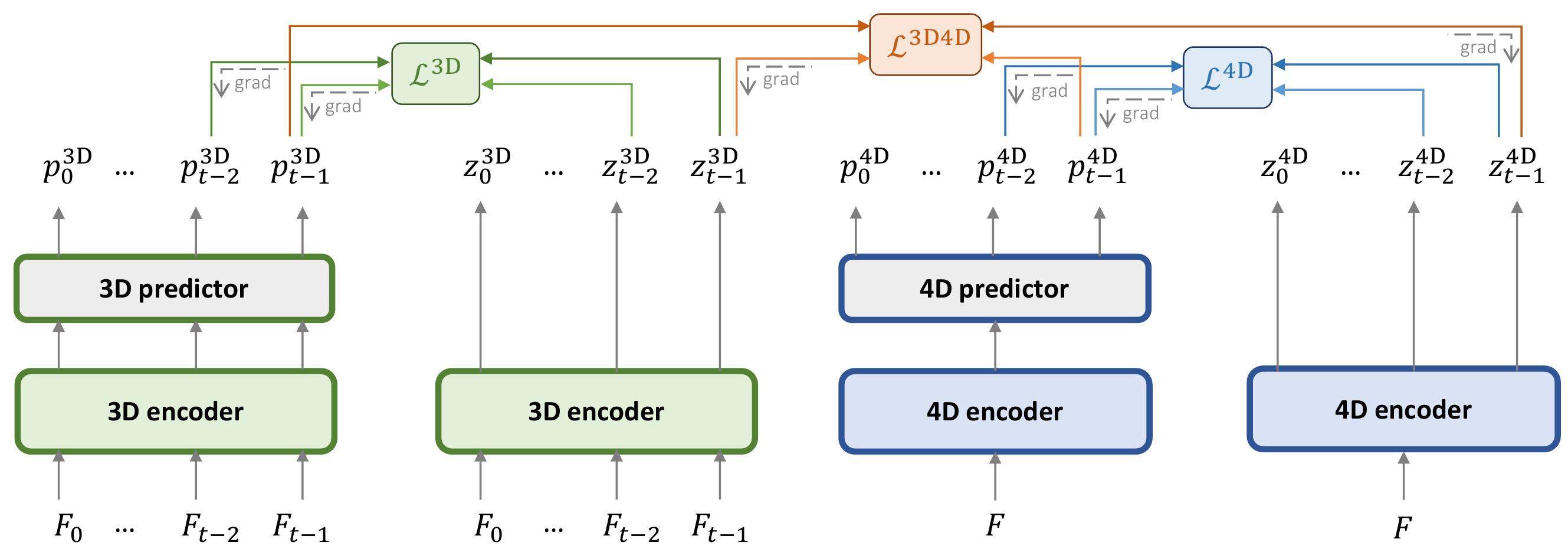}
   \vspace{-0.2cm}
   \caption{4DContrast pre-training. 
   We visualize 3D-3D, 3D-4D, and 4D-4D losses across frame and spatio-temporal correspondence.
   Note that losses are established across all pairs of frames for $\mathcal{L^\textrm{3D}}$ $\mathcal{L^\textrm{4D}}$ and across all frames for $\mathcal{L^\textrm{3D4D}}$; for visualization we only show those associated with frames $F_{t-2}$ and $F_{t-1}$, and only for $F_{t-1}$ for $\mathcal{L^\textrm{3D4D}}$.
   Each loss only propagates back according to the gradient arrows due to stop-gradient operations for stable training.
   }
   \vspace{-0.5cm}
   \label{fig:archi}
\end{figure*}

\subsection{4D-Invariant Contrastive Learning}
\label{subsec:pretext_task}

To imbue effective 4D priors into learned 3D features, we consider a static 3D scan $S$ and a synthetic 3D object $O$ as a train sample, and compose them together to form dynamic object movement in the scene $\{F_0,...,F_{t-1}\}$ for $t$ time steps (as described in Section~\ref{subsec:data_generation}).
We then establish spatial correspondences between frames (3D-3D), spatio-temporal correspondences (3D-4D), and dynamic correspondences (4D-4D) as constraints.
3D features are extracted with a 3D encoder $\Phi_{\textrm{3D}}$ and 4D features with a 4D encoder $\Phi_{\textrm{4D}}$, with respective prediction MLPs $P_{\textrm{3D}}$ and $P_{\textrm{4D}}$.

\noindent \textbf{Inter-Frame Spatial Correspondence.} 
For each pair of frames $(F_i, F_j)$ in a train sequence $F$, we consider their spatial correspondence across sequence frames in order to implicitly pose invariance over the dynamic sequence. 
That is, points that correspond to the same location in the original 3D scene $S$ or original object $O$ should also correspond in feature space.
For the set of corresponding point locations $\mathcal{A}_{i,j}$ from frames $(F_i, F_j)$, we consider each pair of point locations $(\mathbf{a}_i, \mathbf{b}_j)\in\mathcal{A}$, we obtain their 3D backbone features at the respective locations: $p^{\textrm{3D}}_{i,a}=P_{\textrm{3D}}(\Phi_{\textrm{3D}}(F_i))(\mathbf{a}_i)$ and $z^{\textrm{3D}}_{j,b}=\Phi_{\textrm{3D}}(F_j)(\mathbf{b}_j)$.
We then compute a symmetrized negative cosine similarity loss between features of corresponding point locations:
\begin{equation}
\label{equ:sym_cosine_3d}
\begin{split}
\mathcal{L}^{\textrm{3D}}_{\mathcal{A}_{i,j}}  = \sum_{(a,b)\in \mathcal{A}_{i,j}} \left( \frac{1}{2}\mathcal{D}(p^{\textrm{3D}}_{i,a}, SG(z^{\textrm{3D}}_{j,b})) \right. + 
\left. \frac{1}{2}\mathcal{D}(p^{\textrm{3D}}_{j,b}, SG(z^{\textrm{3D}}_{i,a})) \vphantom{\frac{1}{2}} \right).
\end{split}
\end{equation}
In Figure~\ref{fig:archi}, we use \textcolor{green_fig}{green} arrows to indicate constraints between frame $F_{t-2}$ and frame $F_{t-1}$.

We compute Eq.~\ref{equ:sym_cosine_3d} over each pair of frames in the whole sequence $F$:
\begin{equation}
\label{equ:loss_3d}
\mathcal{L}^{\textrm{3D}} = \underset{i<j}{\sum^{t-1}_{i=0}\sum^{t-1}_{j=0}}\mathcal{L}^{\textrm{3D}}_{\mathcal{A}_{i,j}}.
\end{equation}

By establishing constraints across 3D frames in a 4D sequence, we encode pose invariance of moving objects across varying background into the learned 3D features.

\noindent \textbf{Spatio-Temporal Correspondence.}
In addition to implicitly encoding pose invariance of moving objects, we establish explicit 3D-4D correspondences to learn 4D priors, encouraging 4D-invariance in the learned features.
For a train sequence $F=\{F_0,...,F_{t-1}\}$, we use the 4D encoder $\Phi_\textrm{4D}$ and the 4D predictor $P_\textrm{4D}$ to extract 4D features from the whole sequence. 
Then $z^{\textrm{4D}}_{i,a}$ indicates the 4D features output by the 4D encoder $\Phi_\textrm{4D}$ at point location $\mathbf{a}_i$ in frame $i$, and $p^{\textrm{4D}}_{i,a}$ denotes the 4D features output by the 4D predictor $P_\textrm{4D}$.  
Then for a frame $F_i$, we consider each 3D point $\mathbf{a}_i\in \mathcal{A}_i$ in this set of frame points $F_i$, and establish a constraint between its corresponding 3D feature extracted by 3D network ($\Phi_\textrm{3D}$ and $P_\textrm{3D}$) and its corresponding 4D feature extracted by 4D network ($\Phi_\textrm{4D}$ and $P_\textrm{4D}$):
\begin{equation}
\label{equ:sym_cosine_3d4d}
\begin{split}
\mathcal{L}^{\textrm{3D4D}}_{\mathcal{A}_i}  = \sum_{a\in \mathcal{A}_i} \left( \frac{1}{2}\mathcal{D}(SG({p^{\textrm{3D}}_{i,a}}), {z^{\textrm{4D}}_{i,a}}) + \right. 
\left. \frac{1}{2}\mathcal{D}(SG({p^{\textrm{4D}}_{i,a}}), {z^{\textrm{3D}}_{i,a}}) \vphantom{\frac{1}{2}} \right).
\end{split}
\end{equation}
As shown in Figure~\ref{fig:archi}, we use \textcolor{orange_fig}{orange} arrows to indicate constraints of frame $F_{t-1}$.
For the entire input sequence $F$, we calculate Eq.~\ref{equ:sym_cosine_3d4d} for every frame, and the 3D-4D contrastive loss $\mathcal{L}^{\textrm{3D4D}}$ is defined as:
\begin{equation}
\label{equ:loss_3d4d}
\mathcal{L}^{\textrm{3D4D}} = {\sum^{t-1}_{i=0}}\mathcal{L}^{\textrm{3D4D}}_{\mathcal{A}_i}.
\end{equation}

Additionally, in order to learn spatio-temporally consistent 4D representations, we employ 4D-4D correspondence constraints inherent to the 4D features within the same point cloud sequence. 
This is formulated analogously to Eq.~\ref{equ:sym_cosine_3d}, replacing the 3D features with the 4D features from different time steps that correspond spatially:
\begin{equation}
\label{equ:sym_cosine_4d}
\begin{split}
\mathcal{L}^{\textrm{4D}}_{\mathcal{A}_{i,j}}  = \sum_{(a)\in \mathcal{A}_{i,j}} \left( \frac{1}{2}\mathcal{D}(p^{\textrm{4D}}_{i,a}, SG(z^{\textrm{4D}}_{j,a})) \right. + 
\left. \frac{1}{2}\mathcal{D}(p^{\textrm{4D}}_{j,a}, SG(z^{\textrm{4D}}_{i,a})) \vphantom{\frac{1}{2}} \right).
\end{split}
\end{equation}
In Figure~\ref{fig:archi}, we use \textcolor{blue_fig}{blue} arrows to indicate 4D constraints between frame $F_{t-2}$ and frame $F_{t-1}$.
We evaluate Eq.~\ref{equ:sym_cosine_4d} over every pair of frames in the entire input sequence $F$, with the 4D contrastive loss $\mathcal{L}^\textrm{4D}$  defined as:
\begin{equation}
\label{equ:loss_4d}
\mathcal{L}^{\textrm{4D}} = \underset{i<j}{\sum^{t-1}_{i=0}\sum^{t-1}_{j=1}}\mathcal{L}^{\textrm{4D}}_{\mathcal{A}_{i,j}}.
\end{equation}

\noindent\textbf{Joint Learning.} Our overall training loss $\mathcal{L}$ consists of three parts including 3D contrastive loss $\mathcal{L}^{\textrm{3D}}$, 3D-4D contrastive loss $\mathcal{L}^{\textrm{3D4D}}$, and 4D
contrastive loss $\mathcal{L}^{\textrm{4D}}$:
\begin{equation}
\label{equ:loss}
\mathcal{L} = w_{\textrm{3D}}\mathcal{L}^{\textrm{3D}} + w_{\textrm{3D4D}}\mathcal{L}^{\textrm{3D4D}} + w_{\textrm{4D}}\mathcal{L}^{\textrm{4D}},
\end{equation}
where constant weights $w_{\textrm{3D}}$, $w_{\textrm{3D4D}}$ and
$w_{\textrm{4D}}$ are used to balance the losses.

\subsection{Generating 4D Correspondence via Scene-Object Augmentation}
\label{subsec:data_generation}
To learn from 4D sequence data to embed 4D priors into learned 3D representations, we leverage existing large-scale real-world 3D scan datasets in combination with synthetic 3D shape datasets.
This enables generation of 4D correspondences without requiring any labels -- by augmenting static 3D scenes with generated trajectories of a moving synthetic object within the scene, which provides inherent 4D correspondence knowledge across the object motion.
Thus for pre-training, we consider pairs of reconstructed scans and an arbitrarily sampled synthetic 3D shape $(S,O)$, and generate a 4D sequence $F=\{F_0,...,F_{t-1}\}$ by moving the object through the scene.

\noindent\textbf{Trajectory Generation.} 
We first generate a trajectory for $O$ in $S$.
We voxelize $S$ at 10cm voxel resolution, and accumulate occupied surface voxels in the height dimension to acquire a 2D map of the scene geometry.
Valid object locations are then identified as those in the 2D map with a voxel accumulation $\leq 1$, with the max height of the accumulated voxels near to the ground floor (within 20cm of the average floor height).
For the object $O$, we consider all possible 2D locations, and if $O$ does not exceed the valid region (based on its bounding sphere), then the location is taken as a candidate object position.
A random position sampled from these candidate positions is taken as the starting point of the trajectory, we can randomly sample a step distance in $[30,90]$cm and step direction such that the angular change in trajectory is $<150^\circ$, and then select the nearest valid candidate position as the second trajectory point. 
We repeat this process for $t$ time steps in the sequence to obtain 4D scene-object augmentations for pre-training.

\noindent\textbf{4D Sequence Generation.}
A sequence of point clouds are then generated based on the computed object trajectory for the scan, up to sequence length $t$, by  compositing the object into the scene under its translation and rotation steps per frame.
This provides inherent correspondence information between 3D scene locations and 4D object movement through the scene.

\begin{table*}[t]
  \centering
  \caption{Summary of fine-tuning of \OURS{} for various downstream 3D scene understanding tasks and datasets. Our pre-training approach learns effective, transferable features, resulting in notable improvement over  the baseline learning paradigm of training from scratch.
  }
  \vspace{-0.1cm}
  \scalebox{1}{  
  \begin{tabular}{l@{\hskip 0.2in}c@{\hskip 0.3in}c@{\hskip 0.3in}l}
    \toprule
    Datasets & Stats (\#train / \#val) & Task & Gain (from scratch)\\
    \midrule
    \multirow{3}{*}{ScanNetV2~\cite{dai2017scannet}} & \multirow{3}{*}{1.2K / 312 scenes } & Sem. Seg. & \textbf{\textcolor{darkgreen}{+2.3\%}} mIoU\\
    & & Ins. Seg. & \textbf{\textcolor{darkgreen}{+4.2\%}} mAP@0.5\\
    & & Obj. Det. & \textbf{\textcolor{darkgreen}{+5.5\%}} mAP@0.5\\
    \midrule
    SUN RGB-D~\cite{song2015sun} & 5.2K / 5K frames & Obj. Det. & \textbf{\textcolor{darkgreen}{+6.5\%}} mAP@0.5\\
    \bottomrule
  \end{tabular}
  }
  \vspace{-0.3cm}
  \label{tab:overall_down}
\end{table*}

\noindent\textbf{Scene Augmentation.} 
We augment the 4D sequences by randomly sampling different points across the geometry in each individual frame.
We also randomly remove cubic chunks of points in the background 3D scene for additional data variation, with the number of chunks removed randomly sampled from $[5,15]$ and the size of the chunks randomly sampled in $[0.15,0.45]$ as a proportion of the scene extent.
We discard any sequences that do not have enough correspondences in its frames; that is, $\geq 30\%$ of the points in the original scan and $\geq 30\%$ of the points of the synthetic object should be consistently represented in each frame, and each frame must maintain at least 50\% of its points through the augmentation process.
Additionally, we further augment the static 3D frame interpretations of the sequence (but not the sequence) by applying random rotation, translation, and scaling to each individually considered 3D frame.

\subsection{Network Architecture for Pre-training}
During pre-training, we leverage correspondences induced by our 4D data generation, between encoded 3D frames as well as across the encoded 4D sequence.
To this end, we employ 3D and 4D feature extractors as meta-architectures for 4D-invariant learning.

To extract per-point features from a 3D scene, we use a 3D encoder $\Phi_\textrm{3D}$ and a 3D predictor $P_\textrm{3D}$. 
$\Phi_\textrm{3D}$ is a U-Net architecture based on sparse 3D convolutions with residual block followed by a $1\times 1\times 1$ sparse convolutional projection layer, and  $P_\textrm{3D}$ is two $1\times 1\times 1$ sparse convolutional layers.

To extract spatio-temporal features from a 4D sequence, we use a 4D encoder $\Phi_\textrm{4D}$ and a 4D predictor $P_\textrm{4D}$. 
These are structured analogously to the 3D feature extraction, using sparse 4D convolutions instead.
For more detailed architecture specifications, we refer to the supplemental material.

\section{Experimental Setup}
\label{sec:exp_setup}
We demonstrate the effectiveness of our 4D-informed pre-training of learned 3D representations for a variety of downstream 3D scene understanding tasks.

\begin{table}[b]
  \centering
  \vspace{-0.3cm}
  \caption{3D object detection on ScanNet.
  Our \OURS{} pre-training leads to improved performance in comparison with state of the art object detection and 3D pretraining schemes.
  }
  \vspace{-0.1cm}
  \scalebox{0.9}{
  \begin{tabular}{@{\hskip 0.2in}l@{\hskip 0.7in}l@{\hskip 0.7in}l@{\hskip 0.3in}}
    \toprule
    Method & Input & mAP@0.5\\
    \midrule
    DSS~\cite{song2016deep} & Geo + RGB & 6.8\\
    F-PointNet~\cite{qi2018frustum} & Geo + RGB & 10.8\\
    GSPN~\cite{yi2019gspn} & Geo + RGB &  17.7\\
    3D-SIS~\cite{hou20193d} & Geo + RGB & 22.5\\
    VoteNet~\cite{qi2019deep} & Geo + Height & 33.5\\
    \midrule
    \textcolor{gray}{Scratch + VoteNet} & \textcolor{gray}{Geo only} & \textcolor{gray}{34.5} \\
    RandomRooms~\cite{rao2021randomrooms} + VoteNet & Geo only & 36.2~\scriptsize{\textcolor{gray}{(+1.7)}} \\
    PointContrast~\cite{xie2020pointcontrast} + VoteNet & Geo only & 38.0~\scriptsize{\textcolor{gray}{(+3.5)}} \\
    CSC~\cite{hou2021exploring} + VoteNet & Geo only & 39.3~\scriptsize{\textcolor{gray}{(+4.8)}} \\
    Ours + VoteNet & Geo only & \textbf{40.0~\scriptsize{\textcolor{darkgreen}{(+5.5)}}} \\
    \bottomrule
  \end{tabular}
  }
  \vspace{-0.1in}
  \label{tab:det_scannet}
\end{table}

\noindent \textbf{Pre-training Setup.}
We use reconstructed 3D scans from ScanNet~\cite{dai2017scannet} and synthetic 3D shapes from ModelNet~\cite{wu20153d} to compose our 4D sequence data for pre-training.
We use the official ScanNet train split with 1201 train scans, augmented with shapes from ModelNet from eight furniture categories: chair, desk, dresser, nightstand, sofa, table, bathtub, and toilet. 
For each 3D scan, we generate 20 trajectories of an object moving through the scan, following Section~\ref{subsec:data_generation} with $t=4$.
For sequence generation we use 2cm resolution for the scene and 1000 randomly sampled points from the synthetic object to compose together.

The 3D and 4D sparse U-Nets are implemented with MinkowskiEngine~\cite{choy20194d} using 2cm voxel size for 3D and 5cm voxel size for 4D. 
For pre-training we consider only geometry information from the scene-object sequence augmentations.
We use an SGD optimizer with initial learning rate 0.25 and a batch-size of 12. The learning rate is decreased by a factor of 0.99 every 1000 steps. 
We train for 50K steps until convergence. 

\noindent \textbf{Fine-tuning on Downstream Tasks.}
We use the same pre-trained backbone network in the three 3D scene understanding tasks of semantic segmentation, instance segmentation, and object detection.
For semantic segmentation, we directly use the U-Net architecture for dense label prediction, and for object detection and  instance segmentation, we use VoteNet~\cite{qi2019deep} and PointGroup~\cite{jiang2020pointgroup} respectively, both with our pre-trained 3D U-Net backbone.
All experiments, including comparisons with state of the art, are trained with geometric information only, unless otherwise noted.
Fine-tuning experiments on semantic segmentation are trained with a batch size of 48 for 10K steps, using an initial learning rate of 0.8 with polynomial decay with power 0.9.
For instance segmentation, we use the same training setup as PointGroup, and use an initial learning rate of 0.1.
For object detection, the network is trained for 500 epochs, and the learning rate is 0.001 and decayed by a factor of 0.5 at epochs 250, 350, and 450. 
We use a batch size of 6 on ScanNet and 16 on SUN RGB-D.

\section{Results}
We demonstrate that our learned features under 3D-4D constraints can effectively transfer well to a variety of downstream 3D scene understanding tasks.
We consider both in-domain transfer to 3D scene understanding tasks on ScanNet~\cite{dai2017scannet} (Section~\ref{results:scannet}), as well as out-of-domain transfer to SUN RGB-D~\cite{song2015sun} (Section~\ref{results:sunrgbd}); a summary is shown in Table~\ref{tab:overall_down}. We also show data-efficient scene understanding (Section~\ref{results:data-effi}) and additional analysis (Section~\ref{results:analysis}).
Note that for all downstream experiments, we do not use the 4D backbone and thus use the same 3D U-Net architecture as PointContrast~\cite{xie2020pointcontrast} and CSC~\cite{hou2021exploring}.

All experiments, including our method and all baseline comparisons, are trained on geometric data only without any color information.

\subsection{ScanNet}
\label{results:scannet}

We first demonstrate our \OURS{} pre-training in fine-tuning for 3D object detection, semantic segmentation, and instance segmentation on ScanNet~\cite{dai2017scannet}, showing the effectiveness of learning 3D features under 4D constraints.
Tables \ref{tab:det_scannet}, \ref{tab:sem_scannet}, and \ref{tab:ins_scannet} evaluate performance on 3D object detection, semantic segmentation, and instance segmentation, respectively.

Table~\ref{tab:det_scannet} shows 3D object detection results, for which our pretraining approach improves over baseline training from scratch (+5.5\% mAP@0.5) as well as over the strong 3D-based pre-training methods of RandomRooms~\cite{rao2021randomrooms}, PointContrast~\cite{xie2020pointcontrast} and CSC~\cite{hou2021exploring}.

\begin{figure*}[thb]
  \centering
  \vspace{-0.1cm}
   \includegraphics[width=0.97\linewidth]{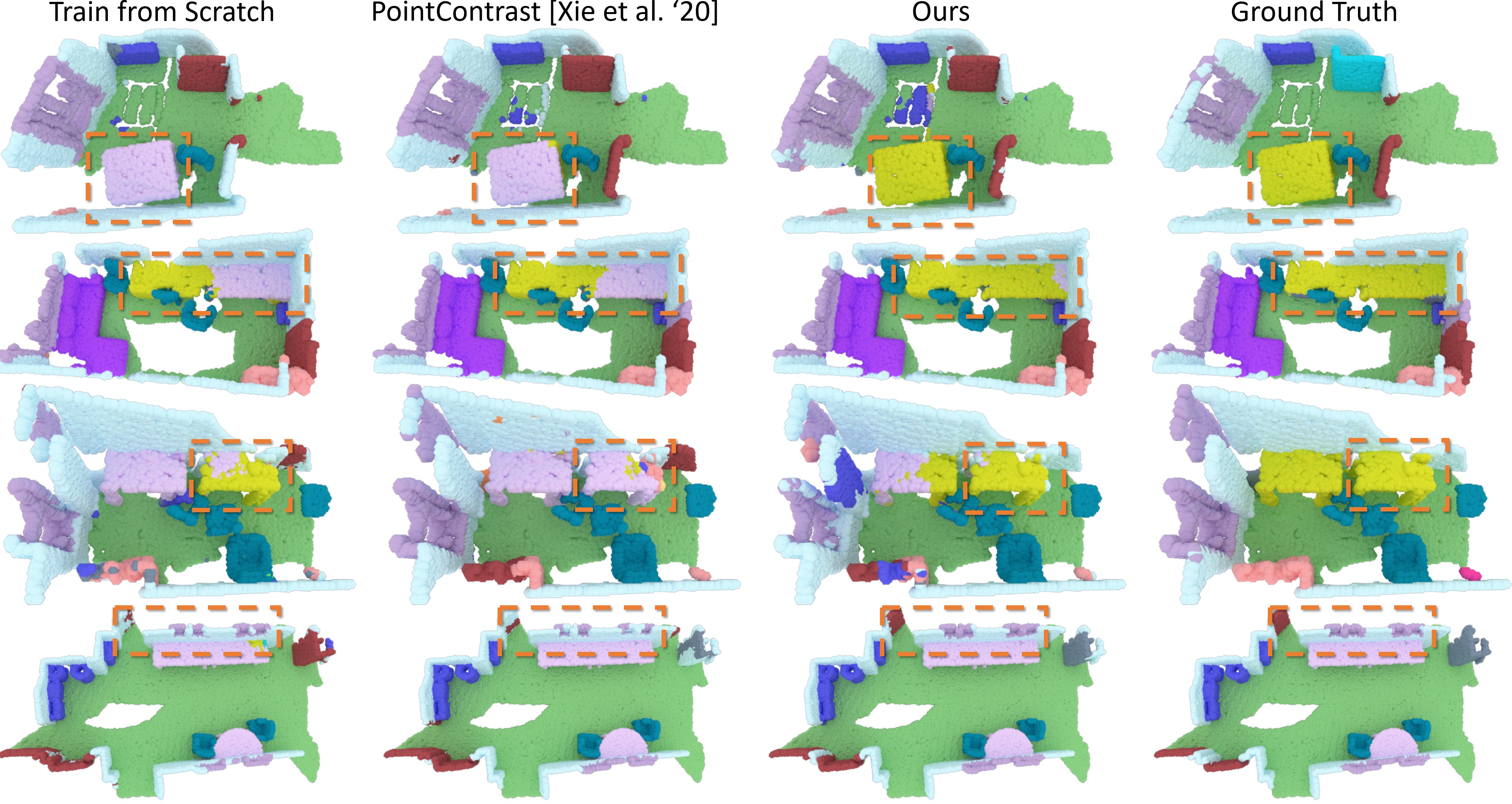}
   \vspace{-0.2cm}
   \caption{Qualitative results on ScanNet semantic segmentation. 
   Our \OURS{} pre-training to encode 4D priors enables more consistent segmentation results, in comparison to training from scratch as well as 3D-based PointContrast~\cite{xie2020pointcontrast} pre-training.
   }
   \vspace{-0.1cm}
   \label{fig:ins_visual}
\end{figure*}

In Tables~\ref{tab:sem_scannet} and \ref{tab:ins_scannet}, we evaluate semantic segmentation in comparison with state-of-the-art 3D pre-training approaches~\cite{hou2021exploring,xie2020pointcontrast}, as well as a baseline training paradigm from scratch.
These pre-training approaches improve notably over training from scratch, and our \OURS{} approach leveraging learned representations under 4D constraints, leads to additional performance improvement over train from scratch (+2.3\% mIoU for semantic segmentation and +4.2\% mAP@0.5 for instance segmentation).
We show qualitative results for semantic segmentation in Figure~\ref{fig:ins_visual}. 

\subsection{SUN RGB-D}
\label{results:sunrgbd}
We additionally show that our \OURS{} learning scheme can produce transferable representations across datasets.
We leverage our pre-trained weights from ScanNet + ModelNet, and explore downstream 3D object detection on the SUN RGB-D~\cite{song2015sun} dataset.
SUN RGB-D is a dataset of RGB-D images, containing 10,335 frames captured with a variety of commodity \mbox{RGB-D} sensors.
It contains 3D object bounding box annotations for 10 class categories. We follow the official train/test split of 5,285 train frames and 5,050 test frames.

\begin{figure}[thbp]
  \centering
   \includegraphics[width=0.9\linewidth]{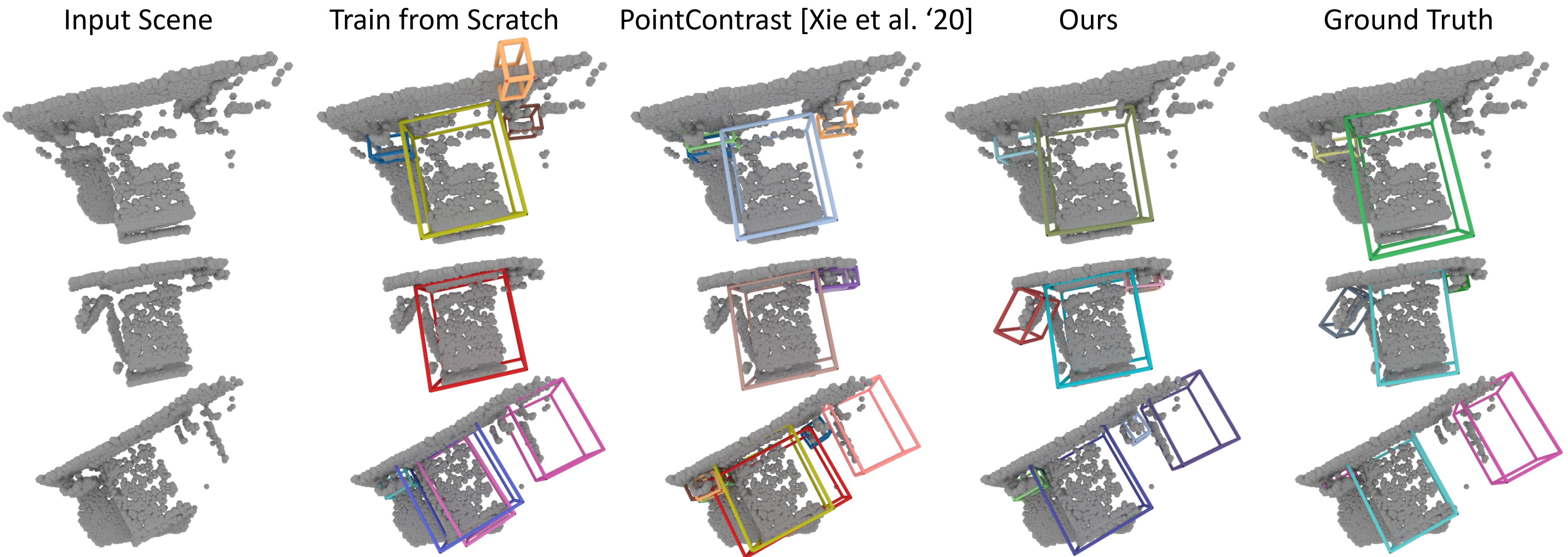}
   \vspace{-0.1cm}
   \caption{ Qualitative results on SUN RGB-D~\cite{song2015sun} object detection. 
   Our \OURS{} pre-training to encode 4D priors enables more accurate detection results, in comparison to training from scratch as well as 3D-based PointContrast~\cite{xie2020pointcontrast} pre-training.
   Different colors denote different objects.
   }
   \label{fig:det_sunrgbd}
\end{figure}

Table~\ref{tab:det_sunrgbd} shows 3D object detection performance on SUN RGB-D, with qualitative results visualized in Figure~\ref{fig:det_sunrgbd}.
We use the same pre-training as with ScanNet, with downstream fine-tuning on SUN RGB-D data.
\OURS{} improves over training from scratch (+6.5\% mAP@0.5), with our learned representations surpassing the 3D-based pre-training \cite{hou2021exploring,rao2021randomrooms,xie2020pointcontrast,Zhang_2021_ICCV}.

\begin{table}[tb]
  \centering
  \caption{Semantic segmentation on ScanNet.
  Our 4D-informed pre-training learns effective features that lead to improved performance boost over training from scratch as well as state-of-the-art 3D-based pre-training of CSC~\cite{hou2021exploring} and PointContrast~\cite{xie2020pointcontrast}.
  }
  \vspace{-0.2cm}
  \scalebox{1}{
  \begin{tabular}{@{\hskip 0.1in}l@{\hskip 0.95in} l@{\hskip 0.95in}l@{\hskip 0.3in}}
    \toprule
     Method & mIoU & mAcc\\
    \midrule
    \textcolor{gray}{Scratch} & \textcolor{gray}{70.0} ~ & \textcolor{gray}{78.1} ~\\
    CSC~\cite{hou2021exploring} & 70.7~\scriptsize{\textcolor{gray}{(+0.7)}} & 78.8~\scriptsize{\textcolor{gray}{(+0.7)}} \\
    PointContrast~\cite{xie2020pointcontrast} & 71.3~\scriptsize{\textcolor{gray}{(+1.3)}} & 79.3~\scriptsize{\textcolor{gray}{(+1.2)}} \\
    Ours &\textbf{72.3~\scriptsize{\textcolor{darkgreen}{(+2.3)}}} & \textbf{80.8~\scriptsize{\textcolor{darkgreen}{(+2.7)}}}\\
    \bottomrule
  \end{tabular}
  }
  \vspace{-0.1cm}
  \label{tab:sem_scannet}
\end{table}

\begin{table}[t]
  \centering
  \caption{Instance segmentation on ScanNet. 
  Our 4D-imbued pre-training leads to significantly improved results over training from scratch, as well as favorable performance over other 3D-only pretraining schemes.
  }
  \vspace{-0.2cm}
  \scalebox{1}{
  \begin{tabular}{@{\hskip 0.1in}l@{\hskip 0.95in} l@{\hskip 0.95in}l@{\hskip 0.3in}}
    \toprule
     Method & mAP@0.5 & mIoU\\
    \midrule
    \textcolor{gray}{Scratch} & \textcolor{gray}{53.4} & \textcolor{gray}{69.0} \\
    PointContrast~\cite{xie2020pointcontrast} & 55.8~\scriptsize{\textcolor{gray}{(+2.4)}}& 70.9~\scriptsize{\textcolor{gray}{(+1.9)}} \\
    CSC~\cite{hou2021exploring} & 56.5~\scriptsize{\textcolor{gray}{(+3.1)}}& 71.1~\scriptsize{\textcolor{gray}{(+2.1)}} \\
    Ours & \textbf{57.6~\scriptsize{\textcolor{darkgreen}{(+4.2)}}} & \textbf{71.4~\scriptsize{\textcolor{darkgreen}{(+2.4)}}} \\
    \bottomrule
  \end{tabular}
  }
  \vspace{-0.1in}
  \label{tab:ins_scannet}
\end{table}

\begin{table}[t]
  \centering
  \caption{3D object detection on SUN RGB-D. Our 4D-based pre-training learns effective 3D representations, improving performance over training from scratch and state-of-the-art 3D pre-training methods.~$^\ast$indicates  that PointNet++ is used as a backbone instead of a 3D U-Net.}
  \vspace{-0.1cm}
  \scalebox{0.93}{
  \begin{tabular}{@{\hskip 0.15in}l@{\hskip 0.7in}l@{\hskip 0.7in}l@{\hskip 0.2in}}
    \toprule
     Method & Input & mAP@0.5 \\
    \midrule
    VoteNet~\cite{qi2019deep} & Geo + Height & 32.9 \\
    \midrule
    \textcolor{gray}{Scratch + VoteNet~\cite{qi2019deep}} & Geo & \textcolor{gray}{31.7} \\
    PointContrast~\cite{xie2020pointcontrast} + VoteNet & Geo & 34.8~\scriptsize{\textcolor{gray}{(+3.1)}} \\
    RandomRooms~\cite{rao2021randomrooms}$^\ast$ + VoteNet & Geo & 35.4~\scriptsize{\textcolor{gray}{(+3.7)}} \\
    DeepContrast~\cite{Zhang_2021_ICCV}$^\ast$ + VoteNet & Geo & 35.5~\scriptsize{\textcolor{gray}{(+3.8)}} \\
    CSC~\cite{hou2021exploring} + VoteNet & Geo & 36.4~\scriptsize{\textcolor{gray}{(+4.7)}} \\
    Ours + VoteNet & Geo & \textbf{38.2~\scriptsize{\textcolor{darkgreen}{(+6.5)}}} \\
    \bottomrule
  \end{tabular}
  }
  \vspace{-0.1in}
  \label{tab:det_sunrgbd}
\end{table}

\begin{figure}[tbh]
    \centering
    \includegraphics[width=0.46\linewidth]{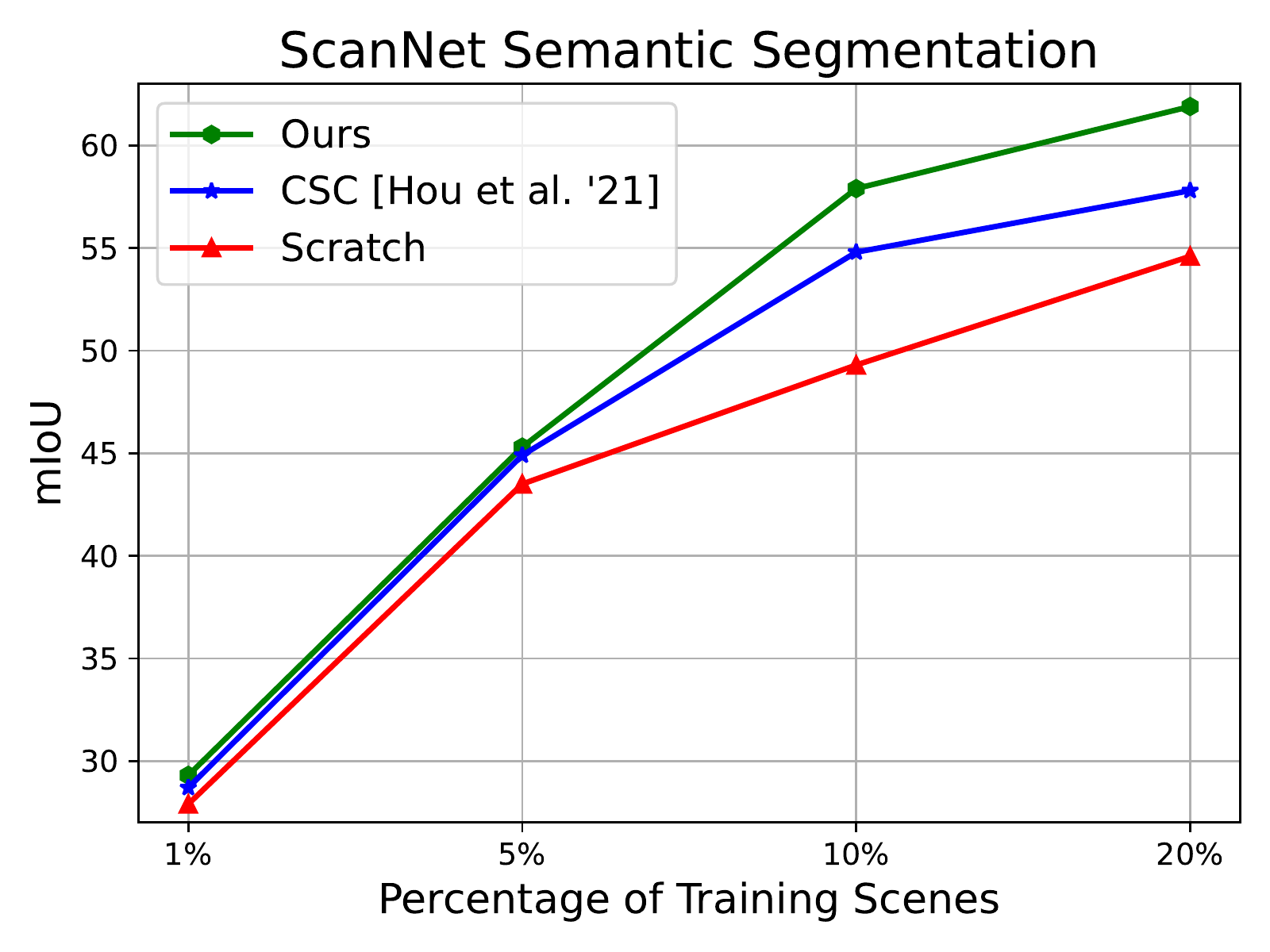}
    \includegraphics[width=0.46\linewidth]{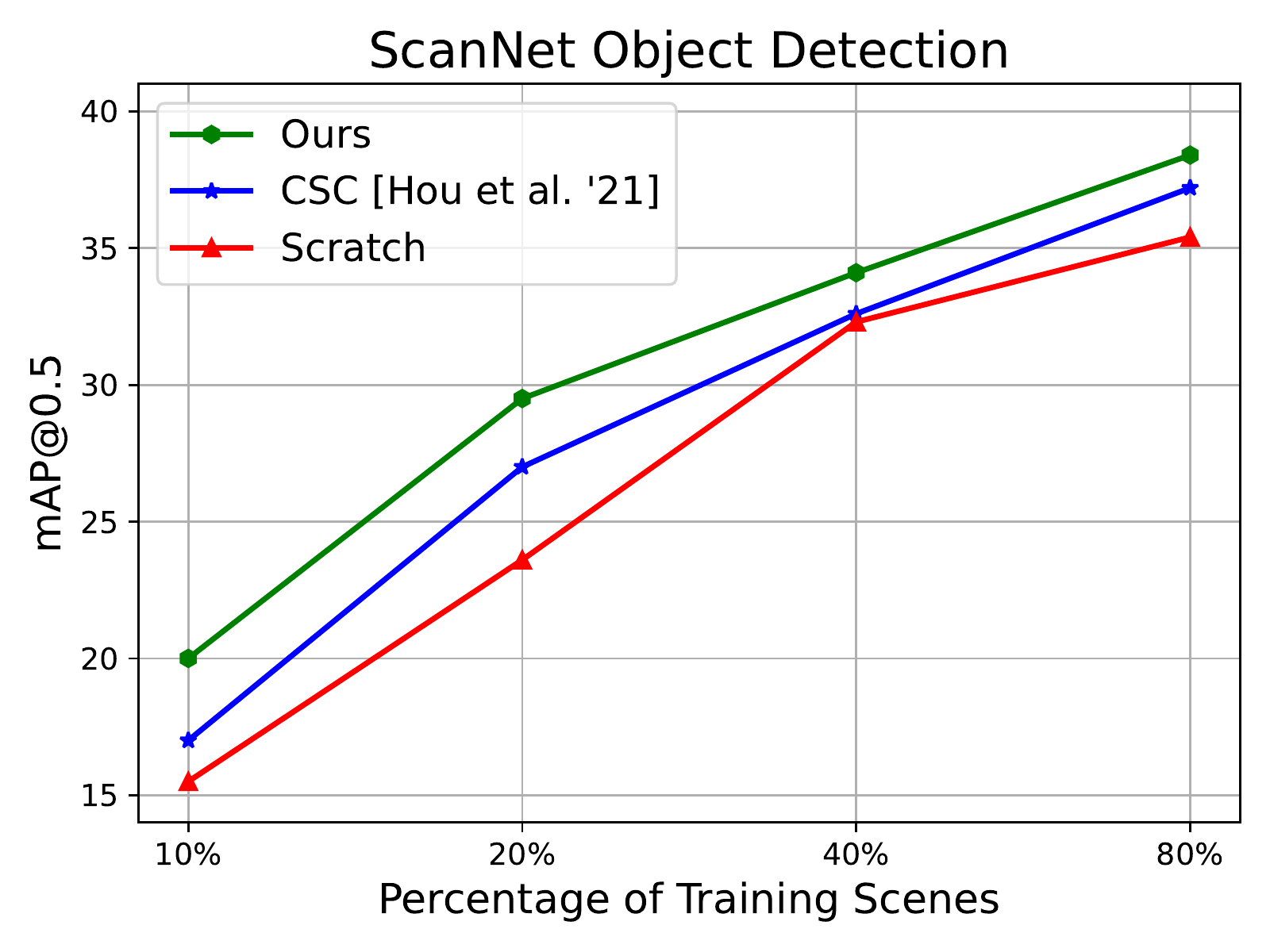}
    \caption{Data-efficient learning on ScanNet semantic segmentation and object detection.
    Under limited data scenarios, our 4D-imbued pre-training effectively improves performance over training from scratch as well as the state-of-the-art CSC~\cite{hou2021exploring}.}
    \label{fig:effi_scannet}
    \vspace{-0.1in}
\end{figure}
\subsection{Data-efficient 3D Scene Understanding}
\label{results:data-effi}
We evaluate our approach in the scenario of limited training data, as shown in Figure~\ref{fig:effi_scannet}. 
\OURS{} improves over baseline training from scratch as well as over state-of-the-art data-efficient scene understanding CSC~\cite{hou2021exploring} in semantic segmentation and object detection under various different percentages of ScanNet training data.
With only 20\% of the training data, we can recover $87\%$ of the fine-tuned semantic segmentation performance training with 100\% of the train data from scratch.
In object detection, our pre-training enables improved performance for all percentage settings, notably in the very limited regime with +3.0/4.5\% mAP@0.5 over CSC/training from scratch at 10\% data, and +2.5/5.9\% mAP@0.5 with 20\% data.

\subsection{Ablation Studies}
\label{results:analysis}
\noindent \textbf{Effect of 3D and 4D Data Augmentation.}  
We consider a baseline variant of our approach that considers only the static 3D scene data without any scene-object augmentations with 3D-3D constraints during pre-training in \mbox{Table~\ref{tab:self_compare}} (\textit{Ours (3D data, 3D-3D only)}), which provides some improvement over training from scratch but is notably improved with our 4D pre-training formulation.
We additionally consider using our 4D scene-object augmentation with only 3D-3D constraints between sequence frames during pre-training (\textit{Ours (4D data, 3D-3D only)}) in Table~\ref{tab:self_compare}, which helps to additionally improve performance with implicitly learned priors from 4D data.
Both are further improved by our approach to explicitly learn 4D priors in 3D features.

\noindent \textbf{Effect of 4D-invariant Contrastive Priors.}
In Table~\ref{tab:self_compare}, we see that learning 4D-invariant contrastive priors through our 3D-4D and 4D-4D constraints during pretraining improves upon data augmentation variants only.
Additionally, Table~\ref{table:oursvs3d} evaluates the 3D variant of our approach with our full 4D-based pre-training across a variety of downstream tasks, showing consistent improvements from learned 4D-based priors. 

\noindent \textbf{Effect of SimSiam Contrastive Learning.} 
We also consider the effect of our SimSiam contrastive framework as PointContrast~\cite{xie2020pointcontrast} leverages a PointInfoNCE contrastive loss.
We note that the 3D variant of our approach (\textit{Ours (3D data, 3D-3D only)}) reflects a  PointContrast~\cite{xie2020pointcontrast} setting using our scene augmentation and SimSiam architecture, which our 4D-based feature learning outperforms.

\begin{table}[tb]
  \centering
  \caption{
  Additionally ablation variants: compared to a baseline of using 3D-3D constraints on static 3D scene data only, leveraging augmented 4D sequence data improves feature learning even under 3D only constraints. 
  Our final \OURS{} pre-training leveraging constraints with learned 4D features achieves the best performance.
  }
  \vspace{-0.15cm}
  \scalebox{0.85}{
  \begin{tabular}{@{\hskip 0.1in}l@{\hskip 0.2in}l@{\hskip 0.2in} l@{\hskip 0.2in}l@{\hskip 0.1in}}
    \toprule
     Method/Data Augmentation & Pre-training Loss Term(s) & mIoU & mAcc\\
    \midrule
    \textcolor{gray}{Scratch} & - & \textcolor{gray}{70.0} ~ & \textcolor{gray}{78.1} ~\\
    Ours (3D data, 3D-3D only) &  $\mathcal{L}^{\textrm{3D}}$(static scene) & 71.5~\scriptsize{\textcolor{gray}{(+1.5)}} &  79.7~\scriptsize{\textcolor{gray}{(+1.6)}}\\
    Ours (4D data, 3D-3D only) & $\mathcal{L}^{\textrm{3D}}$(dynamic scene) & 71.7~\scriptsize{\textcolor{gray}{(+1.7)}} & 80.0~\scriptsize{\textcolor{gray}{(+1.9)}}\\
    Ours & $\mathcal{L}^{\textrm{3D}}$,$\mathcal{L}^{\textrm{3D4D}}$,$\mathcal{L}^{\textrm{4D}}$(dynamic scene) &\textbf{72.3~\scriptsize{\textcolor{darkgreen}{(+2.3)}}} & \textbf{80.5~\scriptsize{\textcolor{darkgreen}{(+2.4)}}}\\
    \bottomrule
  \end{tabular}
  }
  \label{tab:self_compare}
\end{table}

\begin{table}[tb]
\centering
\caption{Extended ablation of the 3d-only variant of our approach on ScanNet.}
\vspace{-0.10cm}
\scalebox{0.85}{
    \begin{tabular}{@{\hskip 0.1in}l@{\hskip 0.7in}c@{\hskip 0.6in}c@{\hskip 0.2in}}
    \toprule
    Task & Ours (3D data, 3D-3D only) & Ours\\
    \midrule
    Ins. Seg. (mAP@0.5) & 54.1 & \textbf{57.6 \scriptsize{\textcolor{darkblue}{(+3.5)}}}\\
    Obj. Det. (mAP@0.5) & 38.7 & \textbf{40.0 \scriptsize{\textcolor{darkblue}{(+1.3)}}} \\
    \midrule
    Sem. Seg. 1\% data (mIoU) & 27.0 & \textbf{28.2 \scriptsize{\textcolor{darkblue}{(+1.2)}}}\\
    Sem. Seg. 5\% data (mIoU) & 44.4 & \textbf{45.3 \scriptsize{\textcolor{darkblue}{(+0.9)}}}\\
    Sem. Seg. 10\% data (mIoU) & 56.3 & \textbf{57.9 \scriptsize{\textcolor{darkblue}{(+1.6)}}}\\
    Sem. Seg. 20\% data (mIoU) & 60.6 & \textbf{61.9 \scriptsize{\textcolor{darkblue}{(+1.3)}}}\\
    Sem. Seg. 100\% data (mIoU) & 71.5 & \textbf{72.3 \scriptsize{\textcolor{darkblue}{(+0.8)}}}\\
    \bottomrule
    \end{tabular}
    }
    \label{table:oursvs3d}
\end{table}

\subsection{Discussion}
While \OURS{} pre-training demonstrates the effectiveness of leveraging 4D priors for learned 3D representations, various limitations remain.
In particular, 4D feature learning with sparse convolutions involves considerable memory during pre-training, so we use half-resolution for characterizing 4D features relative to 3D features and limited sequence durations.
Additionally, we consider a subset of 4D motion when augmenting scenes with moving synthetic objects, and believe exploration of articulated motion or more complex dynamic object interactions would lead to additional insight and robustness of learned feature representations.

\noindent \textit{Memory and Speed.} 
Our 4D-imbued pre-training results in consistent improvements across a variety of tasks and datasets, even with only using the learned 3D backbone for downstream training and inference. 
Thus, our method maintains the same memory and speed costs for inference as purely 3D-based pre-training approaches.
For pre-training, our joint 3D-4D training uses additional parameters (33M for the 4D network in addition to the 38M for the 3D network), but due to jointly learning 4D priors with SimSiam, we do not require as large of a batch size to train as PointContrast~\cite{xie2020pointcontrast} (12 vs their 48), nor as many iterations (up to 30K vs 60K), resulting in slightly less total memory use and pre-training time overall.

\section{Conclusion}

We have presented \OURS{}, a new approach for 3D representation learning that incorporates 4D priors into learned features during pre-training. 
We propose a data augmentation scheme to construct 4D sequences of moving synthetic objects in static 3D scenes, without requiring any semantic labels.
This enables learning from 4D sequences, and we and establish contrastive constraints between learned 3D features and 4D features from the inherent correspondences given in the 4D sequence generation.
Our experiments demonstrate that our 4D-imbued pre-training results in performance improvement across a variety of 3D downstream tasks and datasets.
Additionally, our learned features effectively transfer to limited training data scenarios, significantly outperforming state of the art in the low training data regime.
We hope that this will lead to additional insights in 3D representation learning and new possibilities in 3D scene understanding.

\vspace{0.1in}
\noindent \textbf{Acknowledgements.} This project is funded by the Bavarian State Ministry of Science and the Arts and coordinated by the Bavarian Research Institute for Digital Transformation (bidt), the TUM Institute of Advanced Studies (TUM-IAS), the ERC Starting Grant Scan2CAD (804724), and the German Research Foundation (DFG) Grant Making Machine Learning on Static and Dynamic 3D Data Practical.
\clearpage

%
%
\bibliographystyle{splncs04}
\bibliography{egbib}

\begin{thebibliography}{10}
\providecommand{\url}[1]{\texttt{#1}}
\providecommand{\urlprefix}{URL }
\providecommand{\doi}[1]{https://doi.org/#1}

\bibitem{armeni2017joint}
Armeni, I., Sax, S., Zamir, A.R., Savarese, S.: Joint 2d-3d-semantic data for
  indoor scene understanding. arXiv preprint arXiv:1702.01105  (2017)

\bibitem{chang2017matterport3d}
Chang, A., Dai, A., Funkhouser, T., Halber, M., Niebner, M., Savva, M., Song,
  S., Zeng, A., Zhang, Y.: Matterport3d: Learning from rgb-d data in indoor
  environments. In: International Conference on 3D Vision. pp. 667--676 (2017)

\bibitem{chen2020simple}
Chen, T., Kornblith, S., Norouzi, M., Hinton, G.: A simple framework for
  contrastive learning of visual representations. In: International Conference
  on Machine Learning. pp. 1597--1607 (2020)

\bibitem{chen2020improved}
Chen, X., Fan, H., Girshick, R., He, K.: Improved baselines with momentum
  contrastive learning. arXiv preprint arXiv:2003.04297  (2020)

\bibitem{chen2021exploring}
Chen, X., He, K.: Exploring simple siamese representation learning. In:
  Conference on Computer Vision and Pattern Recognition. pp. 15750--15758
  (2021)

\bibitem{chen2021shape}
Chen, Y., Liu, J., Ni, B., Wang, H., Yang, J., Liu, N., Li, T., Tian, Q.: Shape
  self-correction for unsupervised point cloud understanding. In: International
  Conference on Computer Vision. pp. 8382--8391 (2021)

\bibitem{choy20194d}
Choy, C., Gwak, J., Savarese, S.: 4d spatio-temporal convnets: Minkowski
  convolutional neural networks. In: Conference on Computer Vision and Pattern
  Recognition. pp. 3075--3084 (2019)

\bibitem{choy2019fully}
Choy, C., Park, J., Koltun, V.: Fully convolutional geometric features. In:
  International Conference on Computer Vision. pp. 8958--8966 (2019)

\bibitem{dai2017scannet}
Dai, A., Chang, A.X., Savva, M., Halber, M., Funkhouser, T., Nie{\ss}ner, M.:
  Scannet: Richly-annotated 3d reconstructions of indoor scenes. In: Conference
  on Computer Vision and Pattern Recognition. pp. 5828--5839 (2017)

\bibitem{dai20183dmv}
Dai, A., Nie{\ss}ner, M.: 3dmv: Joint 3d-multi-view prediction for 3d semantic
  scene segmentation. In: European Conference on Computer Vision. pp. 452--468
  (2018)

\bibitem{engelmann20203d}
Engelmann, F., Bokeloh, M., Fathi, A., Leibe, B., Nie{\ss}ner, M.: 3d-mpa:
  Multi-proposal aggregation for 3d semantic instance segmentation. In:
  Conference on Computer Vision and Pattern Recognition. pp. 9031--9040 (2020)

\bibitem{geiger2012we}
Geiger, A., Lenz, P., Urtasun, R.: Are we ready for autonomous driving? the
  kitti vision benchmark suite. In: Conference on Computer Vision and Pattern
  Recognition. pp. 3354--3361 (2012)

\bibitem{graham20183d}
Graham, B., Engelcke, M., Van Der~Maaten, L.: 3d semantic segmentation with
  submanifold sparse convolutional networks. In: Conference on Computer Vision
  and Pattern Recognition. pp. 9224--9232 (2018)

\bibitem{grill2020bootstrap}
Grill, J.B., Strub, F., Altch{\'e}, F., Tallec, C., Richemond, P.H.,
  Buchatskaya, E., Doersch, C., Pires, B.A., Guo, Z.D., Azar, M.G., et~al.:
  Bootstrap your own latent: A new approach to self-supervised learning. arXiv
  preprint arXiv:2006.07733  (2020)

\bibitem{han2020occuseg}
Han, L., Zheng, T., Xu, L., Fang, L.: Occuseg: Occupancy-aware 3d instance
  segmentation. In: Conference on Computer Vision and Pattern Recognition. pp.
  2940--2949 (2020)

\bibitem{hassani2019unsupervised}
Hassani, K., Haley, M.: Unsupervised multi-task feature learning on point
  clouds. In: International Conference on Computer Vision. pp. 8160--8171
  (2019)

\bibitem{he2020momentum}
He, K., Fan, H., Wu, Y., Xie, S., Girshick, R.: Momentum contrast for
  unsupervised visual representation learning. In: Conference on Computer
  Vision and Pattern Recognition. pp. 9729--9738 (2020)

\bibitem{hou20193d}
Hou, J., Dai, A., Nie{\ss}ner, M.: 3d-sis: 3d semantic instance segmentation of
  rgb-d scans. In: Conference on Computer Vision and Pattern Recognition. pp.
  4421--4430 (2019)

\bibitem{hou2021exploring}
Hou, J., Graham, B., Nie{\ss}ner, M., Xie, S.: Exploring data-efficient 3d
  scene understanding with contrastive scene contexts. In: Conference on
  Computer Vision and Pattern Recognition. pp. 15587--15597 (2021)

\bibitem{hu2021bidirectional}
Hu, W., Zhao, H., Jiang, L., Jia, J., Wong, T.T.: Bidirectional projection
  network for cross dimension scene understanding. In: Conference on Computer
  Vision and Pattern Recognition. pp. 14373--14382 (2021)

\bibitem{huang2019texturenet}
Huang, J., Zhang, H., Yi, L., Funkhouser, T., Nie{\ss}ner, M., Guibas, L.J.:
  Texturenet: Consistent local parametrizations for learning from
  high-resolution signals on meshes. In: Conference on Computer Vision and
  Pattern Recognition. pp. 4440--4449 (2019)

\bibitem{huang2021spatio}
Huang, S., Xie, Y., Zhu, S.C., Zhu, Y.: Spatio-temporal self-supervised
  representation learning for 3d point clouds. In: International Conference on
  Computer Vision. pp. 6535--6545 (2021)

\bibitem{jiang2020pointgroup}
Jiang, L., Zhao, H., Shi, S., Liu, S., Fu, C.W., Jia, J.: Pointgroup: Dual-set
  point grouping for 3d instance segmentation. In: Conference on Computer
  Vision and Pattern Recognition. pp. 4867--4876 (2020)

\bibitem{kundu2020virtual}
Kundu, A., Yin, X., Fathi, A., Ross, D., Brewington, B., Funkhouser, T.,
  Pantofaru, C.: Virtual multi-view fusion for 3d semantic segmentation. In:
  European Conference on Computer Vision. pp. 518--535 (2020)

\bibitem{liang2021exploring}
Liang, H., Jiang, C., Feng, D., Chen, X., Xu, H., Liang, X., Zhang, W., Li, Z.,
  Van~Gool, L.: Exploring geometry-aware contrast and clustering harmonization
  for self-supervised 3d object detection. In: International Conference on
  Computer Vision. pp. 3293--3302 (2021)

\bibitem{nekrasov2021mix3d}
Nekrasov, A., Schult, J., Litany, O., Leibe, B., Engelmann, F.: Mix3d:
  Out-of-context data augmentation for 3d scenes. In: 2021 International
  Conference on 3D Vision (3DV). pp. 116--125. IEEE (2021)

\bibitem{nie2021rfd}
Nie, Y., Hou, J., Han, X., Nie{\ss}ner, M.: Rfd-net: Point scene understanding
  by semantic instance reconstruction. In: Conference on Computer Vision and
  Pattern Recognition. pp. 4608--4618 (2021)

\bibitem{qi2020imvotenet}
Qi, C.R., Chen, X., Litany, O., Guibas, L.J.: Imvotenet: Boosting 3d object
  detection in point clouds with image votes. In: Conference on Computer Vision
  and Pattern Recognition. pp. 4404--4413 (2020)

\bibitem{qi2019deep}
Qi, C.R., Litany, O., He, K., Guibas, L.J.: Deep hough voting for 3d object
  detection in point clouds. In: International Conference on Computer Vision.
  pp. 9277--9286 (2019)

\bibitem{qi2018frustum}
Qi, C.R., Liu, W., Wu, C., Su, H., Guibas, L.J.: Frustum pointnets for 3d
  object detection from rgb-d data. In: Conference on Computer Vision and
  Pattern Recognition. pp. 918--927 (2018)

\bibitem{qi2017pointnet}
Qi, C.R., Su, H., Mo, K., Guibas, L.J.: Pointnet: Deep learning on point sets
  for 3d classification and segmentation. In: Conference on Computer Vision and
  Pattern Recognition. pp. 652--660 (2017)

\bibitem{qi2017pointnet++}
Qi, C.R., Yi, L., Su, H., Guibas, L.J.: Pointnet++: Deep hierarchical feature
  learning on point sets in a metric space. In: Neural Information Processing
  Systems (2017)

\bibitem{rao2021randomrooms}
Rao, Y., Liu, B., Wei, Y., Lu, J., Hsieh, C.J., Zhou, J.: Randomrooms:
  Unsupervised pre-training from synthetic shapes and randomized layouts for 3d
  object detection. In: International Conference on Computer Vision. pp.
  3283--3292 (2021)

\bibitem{rozenberszki2022language}
Rozenberszki, D., Litany, O., Dai, A.: Language-grounded indoor 3d semantic
  segmentation in the wild. arXiv preprint arXiv:2204.07761  (2022)

\bibitem{sanghi2020info3d}
Sanghi, A.: Info3d: Representation learning on 3d objects using mutual
  information maximization and contrastive learning. In: European Conference on
  Computer Vision. pp. 626--642 (2020)

\bibitem{sauder2019self}
Sauder, J., Sievers, B.: Self-supervised deep learning on point clouds by
  reconstructing space. Neural Information Processing Systems  (2019)

\bibitem{schult2020dualconvmesh}
Schult, J., Engelmann, F., Kontogianni, T., Leibe, B.: Dualconvmesh-net: Joint
  geodesic and euclidean convolutions on 3d meshes. In: Conference on Computer
  Vision and Pattern Recognition. pp. 8612--8622 (2020)

\bibitem{song2015sun}
Song, S., Lichtenberg, S.P., Xiao, J.: Sun rgb-d: A rgb-d scene understanding
  benchmark suite. In: Conference on Computer Vision and Pattern Recognition.
  pp. 567--576 (2015)

\bibitem{song2016deep}
Song, S., Xiao, J.: Deep sliding shapes for amodal 3d object detection in rgb-d
  images. In: Conference on Computer Vision and Pattern Recognition. pp.
  808--816 (2016)

\bibitem{su2015multi}
Su, H., Maji, S., Kalogerakis, E., Learned-Miller, E.: Multi-view convolutional
  neural networks for 3d shape recognition. In: International Conference on
  Computer Vision. pp. 945--953 (2015)

\bibitem{wang2021unsupervised}
Wang, H., Liu, Q., Yue, X., Lasenby, J., Kusner, M.J.: Unsupervised point cloud
  pre-training via occlusion completion. In: International Conference on
  Computer Vision. pp. 9782--9792 (2021)

\bibitem{wang2021unsupervisedAAAI}
Wang, P.S., Yang, Y.Q., Zou, Q.F., Wu, Z., Liu, Y., Tong, X.: Unsupervised 3d
  learning for shape analysis via multiresolution instance discrimination.
  vol.~35, pp. 2773--2781 (2021)

\bibitem{wu20153d}
Wu, Z., Song, S., Khosla, A., Yu, F., Zhang, L., Tang, X., Xiao, J.: 3d
  shapenets: A deep representation for volumetric shapes. In: Conference on
  Computer Vision and Pattern Recognition. pp. 1912--1920 (2015)

\bibitem{xie2020mlcvnet}
Xie, Q., Lai, Y.K., Wu, J., Wang, Z., Zhang, Y., Xu, K., Wang, J.: Mlcvnet:
  Multi-level context votenet for 3d object detection. In: Conference on
  Computer Vision and Pattern Recognition. pp. 10447--10456 (2020)

\bibitem{xie2020pointcontrast}
Xie, S., Gu, J., Guo, D., Qi, C.R., Guibas, L., Litany, O.: Pointcontrast:
  Unsupervised pre-training for 3d point cloud understanding. In: European
  Conference on Computer Vision. pp. 574--591 (2020)

\bibitem{yi2019gspn}
Yi, L., Zhao, W., Wang, H., Sung, M., Guibas, L.J.: Gspn: Generative shape
  proposal network for 3d instance segmentation in point cloud. In: Conference
  on Computer Vision and Pattern Recognition. pp. 3947--3956 (2019)

\bibitem{zhang2021point}
Zhang, B., Wonka, P.: Point cloud instance segmentation using probabilistic
  embeddings. In: Conference on Computer Vision and Pattern Recognition. pp.
  8883--8892 (2021)

\bibitem{Zhang_2021_ICCV}
Zhang, Z., Girdhar, R., Joulin, A., Misra, I.: Self-supervised pretraining of
  3d features on any point-cloud. In: International Conference on Computer
  Vision. pp. 10252--10263 (2021)

\bibitem{zhang2020h3dnet}
Zhang, Z., Sun, B., Yang, H., Huang, Q.: H3dnet: 3d object detection using
  hybrid geometric primitives. In: European Conference on Computer Vision. pp.
  311--329 (2020)

\end{thebibliography}
\clearpage
\appendix
\setcounter{table}{7}
\setcounter{figure}{6}

\section{Additional Quantitative Analysis}

\begin{table}[tb]
\centering
\caption{Comparisons of alternative 3D backbone and our 4D backbone on ScanNet fine-tuning.}
\vspace{-0.2cm}
\scalebox{1}{
        \begin{tabular}{@{\hskip 0.1in}l@{\hskip 0.4in}c@{\hskip 0.4in}c@{\hskip 0.4in}c@{\hskip 0.13in}}
        \toprule
        Task & \textcolor{gray}{Baseline} & Two 3D Backbones & Ours \\
        \midrule
        Sem.Seg (mIoU)  & \textcolor{gray}{70.0}  & 71.8~\scriptsize{\textcolor{gray}{(+1.8)}} & \textbf{72.3~\scriptsize{\textcolor{darkgreen}{(+2.3)}}} \\
        \midrule
        Ins.Seg (mAP@0.5)  & \textcolor{gray}{53.4}  & 56.2~\scriptsize{\textcolor{gray}{(+2.8)}} & \textbf{57.6~\scriptsize{\textcolor{darkgreen}{(+4.2)}}} \\
        \bottomrule
        \end{tabular}
        }
        \label{table:3d-4d}
\end{table}

\begin{table}[tb]
\centering
\caption{Effect of sequence length of pre-training dynamic data on ScanNet semantic segmentation fine-tuning. A sequence length of 4 helps 4DContrast get higher semantic segmentation mIoU.}
\vspace{-0.2cm}
\scalebox{1}{
    \begin{tabular}{@{\hskip 0.2in}c@{\hskip 0.8in}c@{\hskip 0.8in}c@{\hskip 0.8in}c@{\hskip 0.4in}}
    \toprule
    Sequence Length & 3 & 4 & 5\\
    \midrule
    mIoU & 71.9 & \textbf{72.3} & 71.0\\
    \bottomrule
    \end{tabular}
    }
    \label{table:seq_length}
\end{table}

\noindent \textbf{Alternative 3D Backbone vs 4D Pre-training.}
Our 4D pre-training can help to learn objectness priors from dynamic object movement, in contrast to multiple 3D backbones.
To demonstrate this, we pre-trained with an additional 3D backbone (comparable to our 4D backbone in parameters and UNet structure); this resulted in worse performance than our 4D pre-training which has +0.5 mIoU and +1.4 mAP@0.5 vs. multiple 3D backbones in the tasks of 3D semantic and instance segmentation on ScanNet (as shown in Table~\ref{table:3d-4d}).

\noindent\textbf{Sequence Length Ablation.}
We study the effect of the sequence length of the generated dynamic data used for pre-training in Table~\ref{table:seq_length}.
We consider sequences of length 3, 4, or 5, and set the batch size (number of sequences) to 16, 12, and 10, respectively, to balance the scene frames in each batch during pre-training. 
We find a sequence length of 4 results in more effective feature learning for downstream tasks.

\begin{table}[tbh]
\centering
\caption{Comparisons of SimCLR and SimSiam as our contrastive learning framework on ScanNet semantic segmentation fine-tuning.}
\vspace{-0.2cm}
\scalebox{1}{
        \begin{tabular}{@{\hskip 0.1in}c@{\hskip 0.3in}c@{\hskip 0.3in}c@{\hskip 0.3in}c@{\hskip 0.1in}}
        \toprule
        Contrastive Framework & \textcolor{gray}{Baseline} & Ours (SimCLR) & Ours (SimSiam) \\
        \midrule
        mIoU  & \textcolor{gray}{70.0}  & 71.6~\scriptsize{\textcolor{gray}{(+1.6)}} & \textbf{72.3~\scriptsize{\textcolor{darkgreen}{(+2.3)}}} \\
        \bottomrule
        \end{tabular}
        }
        \label{table:cl_frameworks}
\end{table}

\begin{table}[ht]
\centering
\caption{3D object detection on ScanNet with H3DNet.}
\vspace{-0.2cm}
\scalebox{1}{
        \begin{tabular}{@{\hskip 0.4in}l@{\hskip 1.9in}l@{\hskip 0.7in}}
        \toprule
        Method & mAP@0.5\\
        \midrule
        H3DNet & 43.4 \\
        Ours + H3DNet & \textbf{47.7 \scriptsize{\textcolor{darkgreen}{(+4.3)}}}\\
        \bottomrule
        \end{tabular}
        }
        \label{table:h3dnet}
\end{table}

\begin{table}[th]
\centering
\caption{Semantic segmentation on ScanNet val with Mix3D.}
\vspace{-0.2cm}
\scalebox{1}{
        \begin{tabular}{@{\hskip 0.4in}l@{\hskip 1.4in}l@{\hskip 0.4in}}
        \toprule
        Method & mIoU \\
        \midrule
        Mix3D + MinkowskiNet & 73.9\\
        Ours + Mix3D + MinkowskiNet & \textbf{74.6 \scriptsize{\textcolor{darkgreen}{(+0.7)}}}\\
        \bottomrule
        \end{tabular}
        }
        \label{table:mix3d}
\end{table}

\begin{table}[th]
\centering
\caption{3D classification on ModelNet with Mix3D.}
\vspace{-0.2cm}
\scalebox{1}{
        \begin{tabular}{@{\hskip 0.3in}l@{\hskip 0.6in}c@{\hskip 0.9in}l@{\hskip 0.4in}}
        \toprule
        Method & Voxel Size & Acc\\
        \midrule
        MinkowskiNet & 0.05 & 86.1 \\
        Ours + MinkowskiNet & 0.05 & \textbf{88.5 \scriptsize{\textcolor{darkgreen}{(+2.4)}}}\\
        \midrule
        MinkowskiNet & 0.02 & 90.7 \\
        Ours + MinkowskiNet & 0.02 & \textbf{91.8 \scriptsize{\textcolor{darkgreen}{(+1.1)}}}\\
        \bottomrule
        \end{tabular}
        }
        \label{table:modelnet}
\end{table}

\begin{table}[th!]
\centering
\caption{Semantic and instance segmentation on S3DIS.}
\vspace{-0.2cm}
\scalebox{1}{
        \begin{tabular}{@{\hskip 0.4in}l@{\hskip 0.9in}c@{\hskip 0.9in}c@{\hskip 0.3in}}
        \toprule
        Task & scratch & Ours\\
        \midrule
        Sem.Seg. (mIoU) & 58.6 & \textbf{61.0 \scriptsize{\textcolor{darkgreen}{(+2.4)}}}\\
        \midrule
        Ins.Seg. (mAP@0.5) & 45.8 & \textbf{53.2 \scriptsize{\textcolor{darkgreen}{(+7.4)}}}\\
        \bottomrule
        \end{tabular}
        }
        \label{table:s3dis}
\end{table}

\begin{figure*}[h]
    \includegraphics[width=1\linewidth]{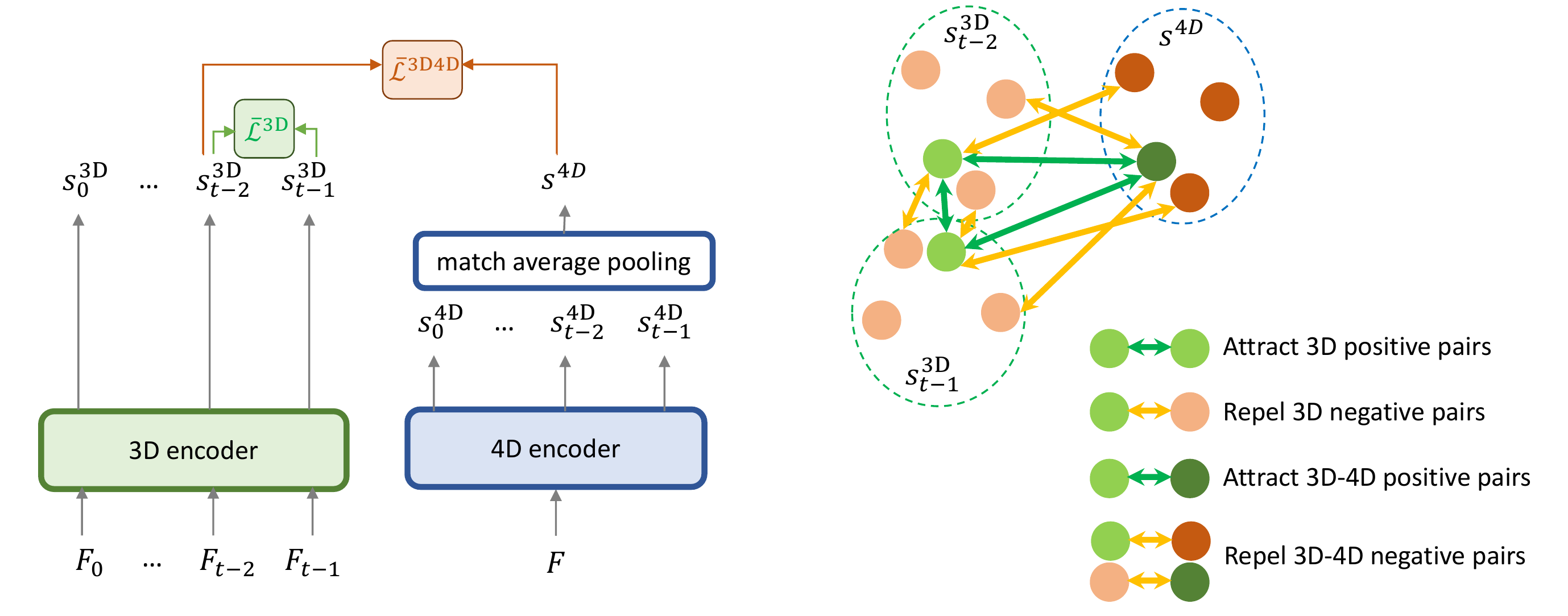}
    \caption{Network architectures of our method using SimCLR as the contrastive learning framework. \textbf{Left:} we show 3D-3D and 4D-4D losses across frame and spatio-temporal correspondence. We only visualize the inter-frame correspondence for $F_{t-2}$ and $F_{t-1}$, and only spatio-temporal correspondence for for $F_{t-2}$, while those loss are established across all pairs of frames for $\mathcal{\bar{L}^\textrm{3D}}$ and all frames for $\mathcal{\bar{L}^\textrm{3D4D}}$.
    \textbf{Right:} we visualize the contrastive losses between 3D feautres of $F_{t-2}$ and $F_{t-1}$ and the 4D feature after match average pooling. The positive pairs is same with Section 3.2 and the negative losses is only calculated for the hardest negative pairs.
    }
    \label{fig:simclr_archi}
\end{figure*}

\noindent\textbf{Comparison of Different Contrastive Frameworks.}
As analyzed in Section~3.1 of the main paper, SimSiam~\cite{chen2021exploring} enables contrastive learning without requiring negative samples or large batch size. We thus verify how these attributes fit our high-dimensional pre-training design by comparing SimSiam and SimCLR~\cite{chen2020simple} as our contrastive framework. 
As shown in Figure~\ref{fig:simclr_archi}, Ours~(SimCLR) removes the 3D and 4D predictors from Ours~(SimSiam), and uses a match average pooling to average 4D features in different frames according to spatial correspondences. For each pair of frames ($F_i,F_j$) in a train sequence, we apply a 3D contrastive loss  $\mathcal{\bar{L}^\textrm{3D}}$ as $\mathcal{{L}^\textrm{3D}}$ (Eq.~4 in Section~3.2). Similar to $\mathcal{{L}^\textrm{3D4D}}$ (Eq.~6 in Section~3.2), we use a 3D-4D contrastive loss $\mathcal{\bar{L}^\textrm{3D4D}}$ to establish correspondence between 3D features and the averaged 4D features. The Hardest-Contrastive loss is borrowed from FCGF \cite{choy2019fully} and PointContrast \cite{xie2020pointcontrast}. 
Note that in our implementation, we find that the PointInfoNCE loss \cite{xie2020pointcontrast} is not as stable as Hardest-Contrastive loss, likely due to  different data augmentation methods between Ours~(SimCLR) and PointContrast.
As shown in Table~\ref{table:cl_frameworks}, \OURS{} coupled with SimSiam framework more effectively leverages the learned representations for improved semantic segmentation performance on ScanNet.

\noindent\textbf{H3DNet Object Detection with \OURS{}.} In Table~\ref{table:h3dnet}, we apply our pre-trained weights to H3DNet \cite{zhang2020h3dnet} (1 descriptor computation tower of its backbone architecture). \OURS{} surpasses training from scratch by 4.3 mAP@0.5 on ScanNet.

\noindent\textbf{Mix3D Semantic Segmentation with \OURS{}.} While \OURS{} focuses on imbuing 4D priors during pre-training to provide effective features for a variety of downstream tasks, Mix3D \cite{nekrasov2021mix3d} tackles a complementary problem of data augmentation during training. 
As shown in Table~\ref{table:mix3d}, our pre-training can be used together with Mix3D to further improve semantic segmentation performance on ScanNet (geometry only input).

\noindent\textbf{MinkowskiNet 3D Classification with \OURS{}.}  We evaluate ModelNet classification accuracy in comparison with MinkowskiNet \cite{choy20194d} trained from scratch for various voxel sizes in Table~\ref{table:modelnet}. Our pre-training shows consistent improvements in both settings.

\noindent\textbf{S3DIS dataset.} We finetune our pre-trained weights for S3DIS \cite{armeni2017joint} segmentation (geometry-only), and consistently improve over training from scratch (as shown in Table~\ref{table:s3dis}): we achieve +2.4 mIoU in semantic segmentation (61.0 ours vs 58.6 scratch) and +7.4 mAP@0.5 in instance segmentation (53.2 vs 45.8).

\section{Network Architecture Details}
\begin{figure*}
    \centering
    \includegraphics[width=1\linewidth]{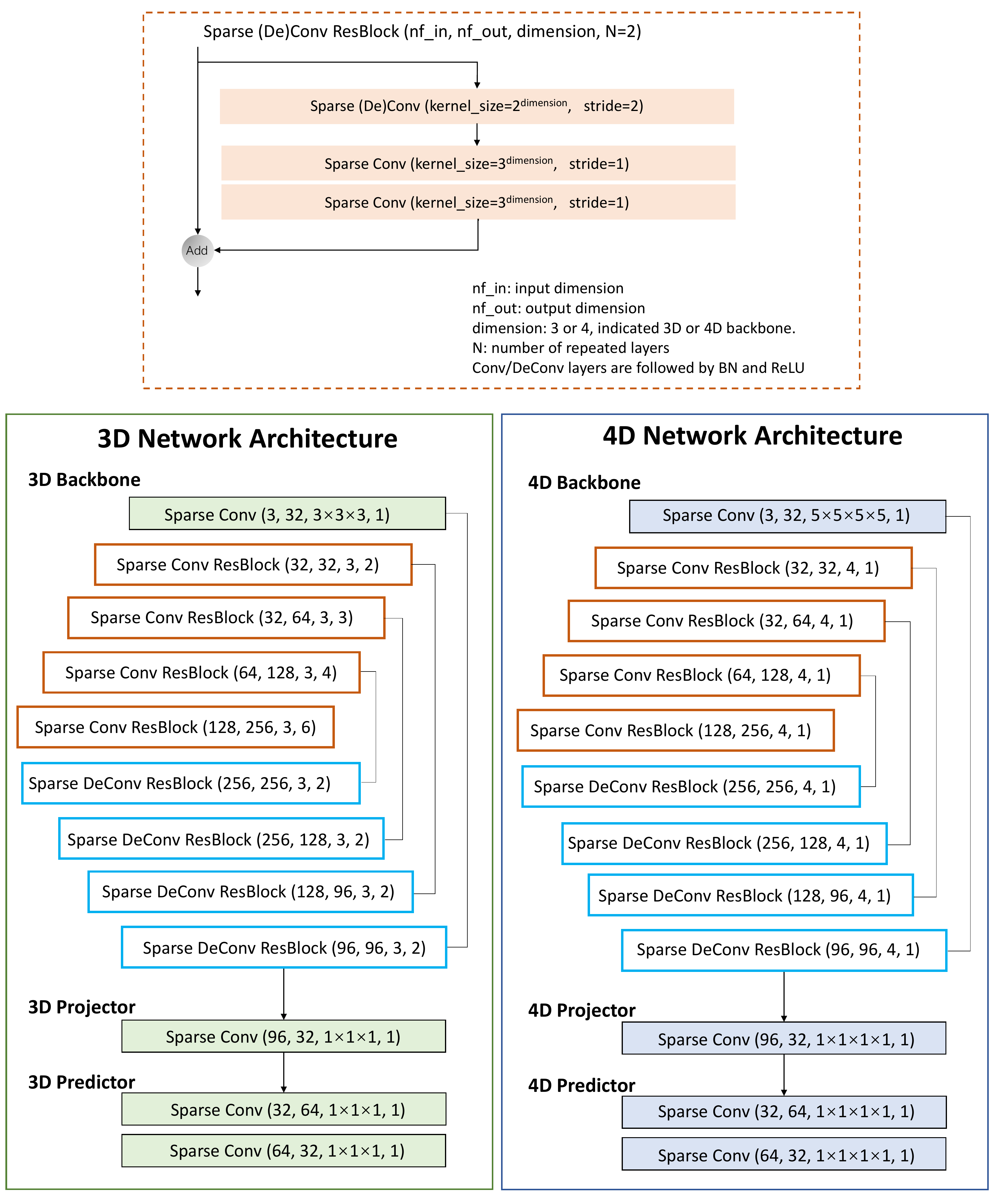}
    \caption{Network architectures of 4DContrast for pre-training. For downstream fine-tuning, only the 3D backbone is kept and fine-tuned.}
    \label{fig:supp_archi}
\end{figure*}

Figure~\ref{fig:supp_archi} details our network architectures. The backbones are a U-Net architecture with sparse convolutions \cite{choy20194d}. We use a 34-layer U-Net as the 3D backbone and a 14-layer U-Net as the 4D backbone. For the 3D and 4D projectors, we use a one-layer sparse convolutional layer with kernel size as $1\times1\times1$ and  $1\times1\times1\times1$, respectively. For the 3D and 4D predictor, we use two sparse convolutional layers. We repeat occupancy into 3-dimension to fit the network input dimension of 3.

\section{Visualization of the Generated 4D Data}
\begin{figure*}
    \centering
    \includegraphics[width=1\linewidth]{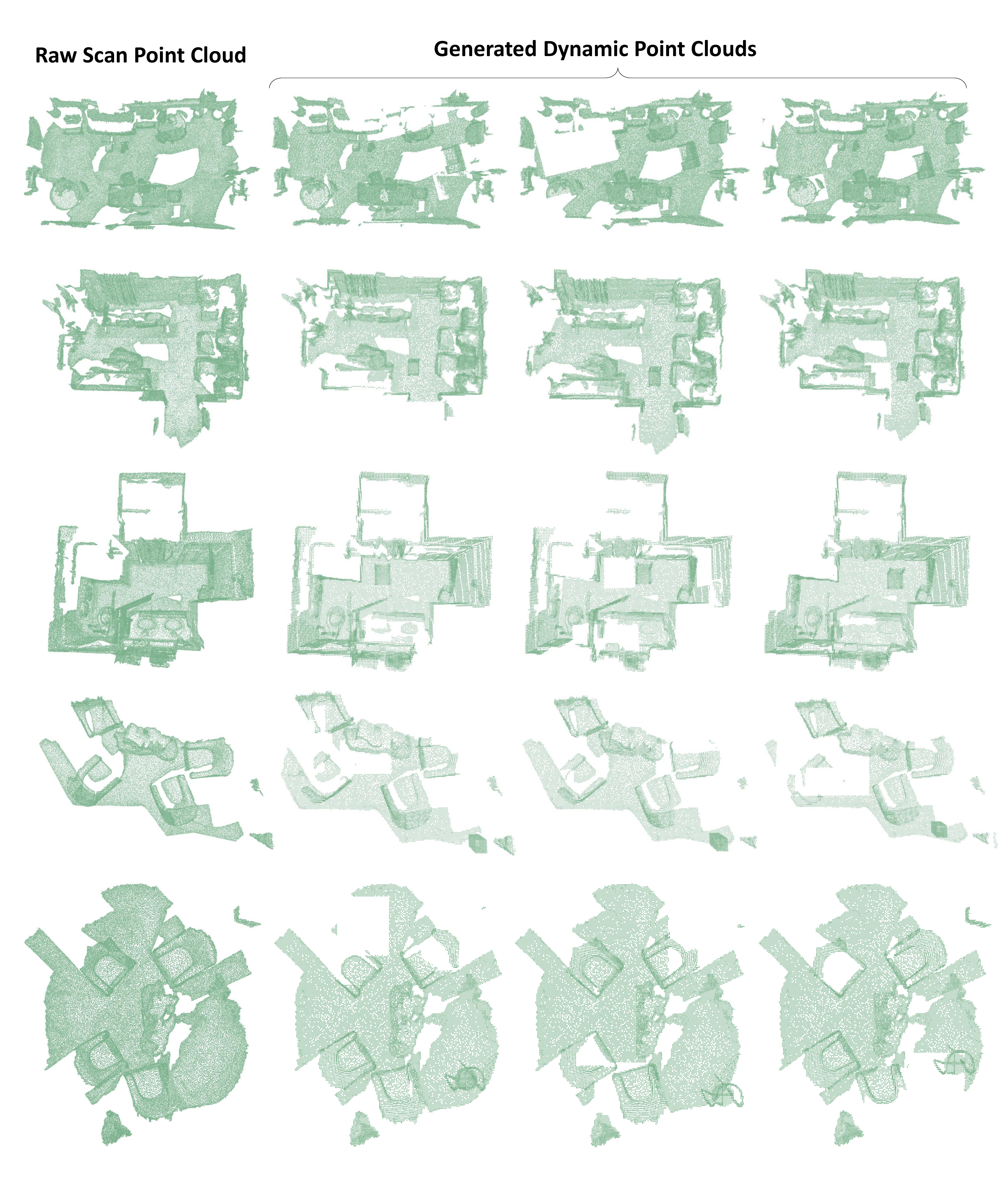}
    \caption{Visualization of generated 4D sequence data. Each row corresponds to a sampled scene. From left to right: raw scan mesh vertices as input cloud, generated dynamic point clouds with scene augmentation and object motion (in three frames). 
    }
    \label{fig:4d_data}
\end{figure*}

Figure~\ref{fig:4d_data} shows the generated 4D data by scene-object augmentation (as described in Section 3.3 of the main paper).

\clearpage
\end{document}